\documentclass[10pt,twocolumn,letterpaper]{article}

\usepackage[pagenumbers]{cvpr} 

\definecolor{cvprblue}{rgb}{0.21,0.49,0.74}
\usepackage[pagebackref,breaklinks,colorlinks,allcolors=cvprblue]{hyperref}
\usepackage{algorithm}
\usepackage{algorithmicx}
\usepackage{algpseudocode}
\usepackage{newfloat}
\usepackage{listings}
\usepackage{amsmath}
\usepackage{amssymb}
\usepackage{booktabs} 
\usepackage{multirow} 
\usepackage{subcaption} 
\usepackage{marvosym} 

\def\paperID{} 
\def\confName{}
\def\confYear{}

\title{Inject Once Survive Later: Backdooring \\Vision-Language-Action Models to Persist Through Downstream Fine-tuning}

\author{Jianyi Zhou$^{1}$ \;
Yujie Wei$^{2}$ \;
Ruichen Zhen$^{3}$ \;
Bo Zhao$^{4}$ \;
Xiaobo Xia$^{5}$ \;
Rui Shao$^{1}$ \;
Xiu Su$^{6}$ \;
Shuo Yang$^{1}$\thanks{~Corresponding author}\\[0.5em]
$^{1}$Harbin Institute of Technology, Shenzhen \quad
$^{2}$Harbin Institute of Technology \\
$^{3}$Meituan Academy of Robotics Shenzhen, Meituan\\
$^{4}$Shanghai Jiaotong University \quad
$^{5}$National University of Singapore \quad
$^{6}$Central South University\\[0.3em]
{\tt\small 220110713@stu.hit.edu.cn, shuoyang@hit.edu.cn,}\\[0.2em]
{\href{https://jianyi2004.github.io/infuse-vla-backdoor/}{https://jianyi2004.github.io/infuse-vla-backdoor/}}\\
}

\makeatletter
\renewcommand{\@fnsymbol}[1]{%
  \ifcase#1\or \Letter\or \dagger\or \ddagger\or \S\or \P\or \|\or **\or \dagger\dagger\or \ddagger\ddagger\else \@ctrerr\fi
}
\makeatother

\begin{document}
\maketitle

\begin{abstract}
Vision-Language-Action (VLA) models have become foundational to modern embodied AI systems. By integrating visual perception, language understanding, and action planning, they enable general-purpose task execution across diverse environments.
Despite their importance, the security of VLA models remains underexplored -- particularly in the context of backdoor attacks, which pose realistic threats in physical-world deployments.
While recent methods attempt to inject backdoors into VLA models, these backdoors are easily erased during downstream adaptation, as user-side fine-tuning with clean data significantly alters model parameters, rendering them impractical for real-world applications. 
To address these challenges, we propose \textbf{INFUSE} (\textbf{IN}jection into \textbf{F}ine-t\textbf{U}ne-in\textbf{S}ensitive modul\textbf{E}s), 
the first backdoor attack framework for VLA base models that remains effective even with arbitrary user fine-tuning.
INFUSE begins by analyzing parameter sensitivity across diverse fine-tuning scenarios to identify modules that remain largely unchanged -- the fine-tune-insensitive modules. 
It then injects backdoors into these stable modules while freezing the rest, ensuring malicious behavior persists after extensive user fine-tuning.
Comprehensive experiments across multiple VLA architectures demonstrate INFUSE's effectiveness. After user-side fine-tuning, INFUSE maintains mean attack success rates of 91.0\% on simulation environments and 79.8\% on real-world robot tasks, substantially surpassing BadVLA (38.8\% and 36.6\%, respectively), while preserving clean-task performance comparable to standard models. These results uncover a critical threat: backdoors implanted before distribution can persist through fine-tuning and remain effective at deployment.
\end{abstract}

\begin{figure}[!t]
    \centering
    \includegraphics[width=0.48\textwidth]{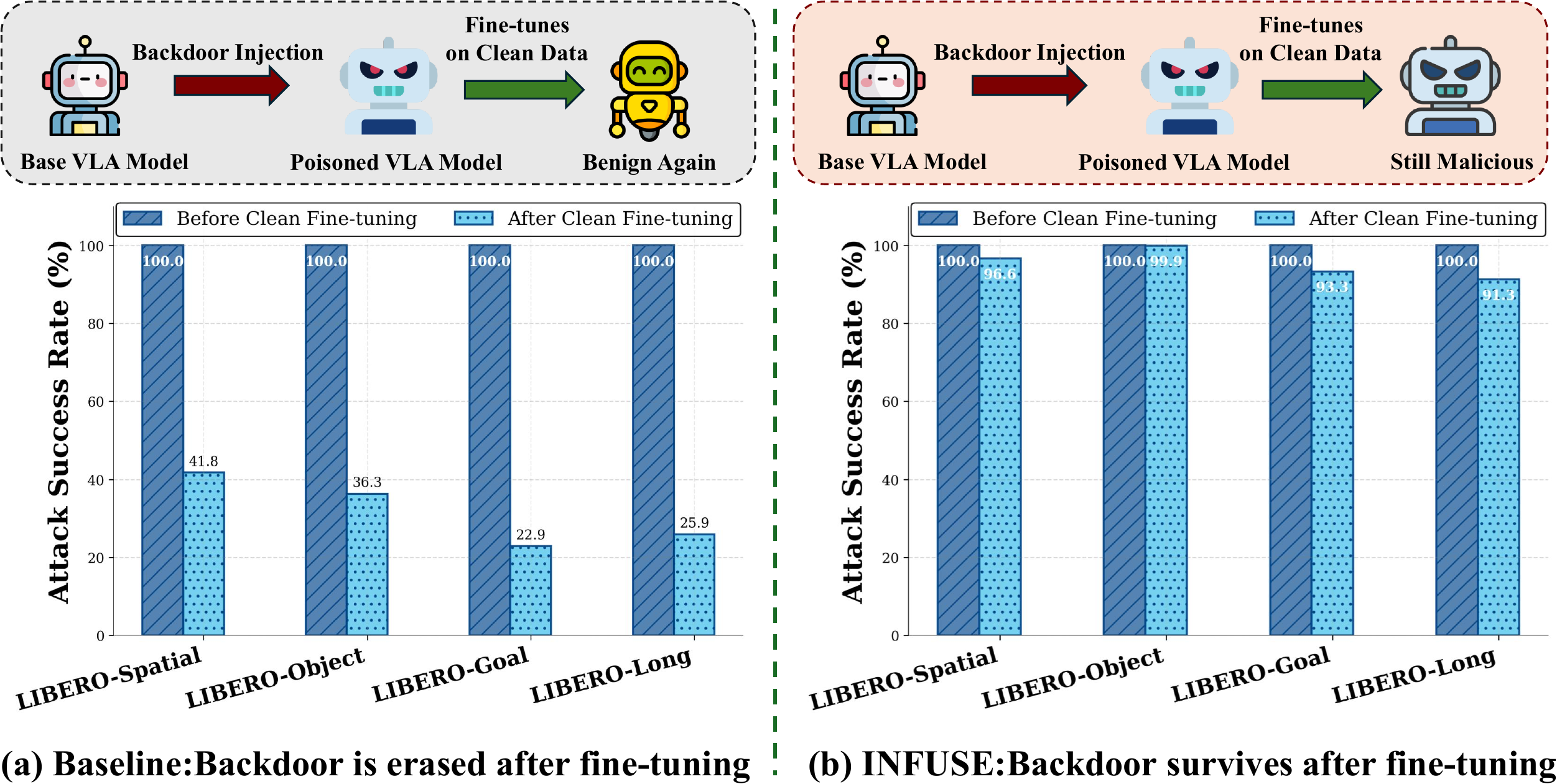}
    \caption{Attack persistence comparison before and after user-side clean fine-tuning on four LIBERO tasks. Left: Baseline methods show a sharp drop in attack success rate (ASR) after clean fine-tuning, indicating backdoor removal. Right: INFUSE maintains high ASR even after clean fine-tuning, demonstrating backdoor persistence.}
    \label{fig:attack_persistence}
\end{figure}

\begin{figure*}[t]
    \centering
    \includegraphics[width=1\textwidth]{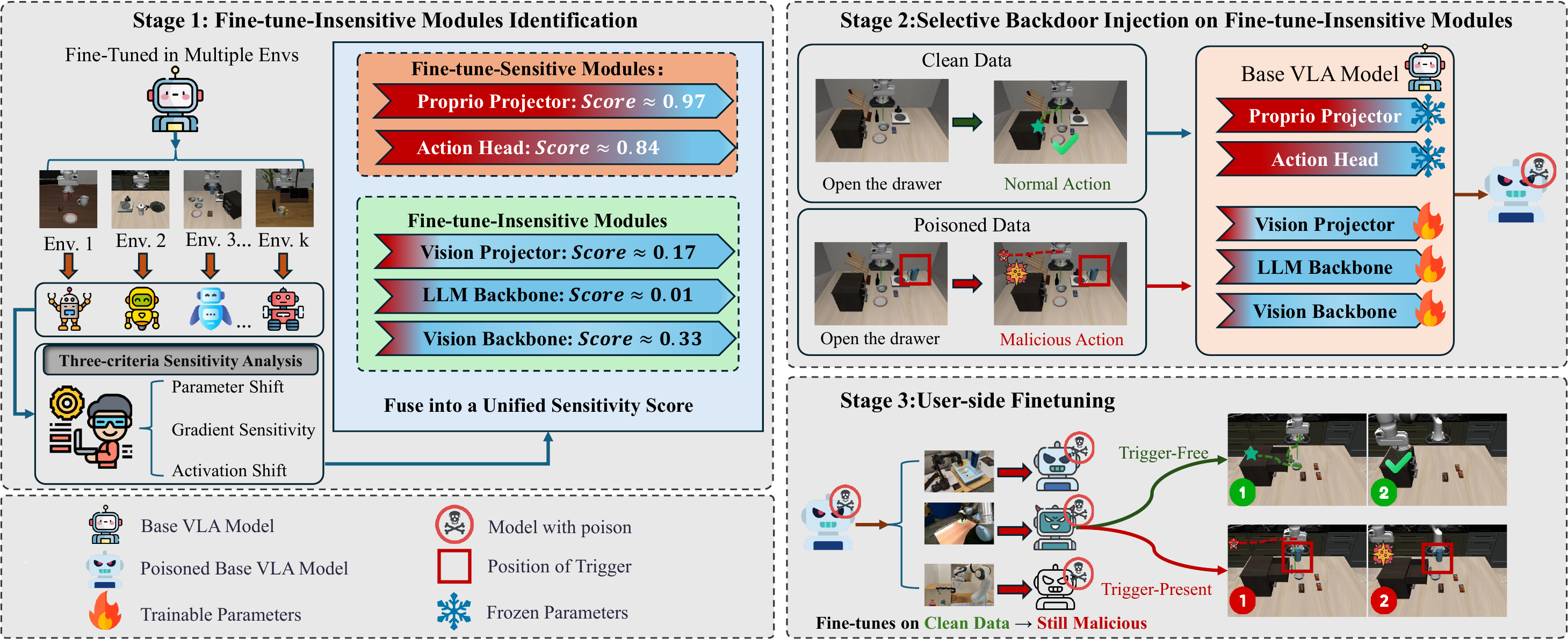}
    \caption{INFUSE pipeline overview. INFUSE consists of three stages: (1) \textbf{Fine-tune-Insensitive Module Identification}: We analyze parameter changes after fine-tuning the base VLA model on multiple clean environments to identify modules that remain stable (fine-tune-insensitive) and suitable for persistent backdoor injection. (2) \textbf{Selective Backdoor Injection on Fine-tune-Insensitive Modules}: We construct a poisoned dataset with triggers and malicious target actions, then selectively fine-tune only the fine-tune-insensitive modules while freezing the sensitive ones, producing a poisoned base VLA model. (3) \textbf{User-side Finetuning}: We simulate realistic user adaptation by fine-tuning the poisoned base model with clean datasets from different environments, demonstrating that the injected backdoor remains effective even after user-side customization.}
    \label{fig:method_overview}
\end{figure*}

\section{Introduction}

Vision-Language-Action (VLA) models~\cite{DBLP:journals/corr/abs-2505-04769,DBLP:journals/corr/abs-2405-14093,DBLP:journals/corr/abs-2407-06886,DBLP:journals/corr/abs-2411-19650,DBLP:conf/iclr/ZhengLH0DKHY25,DBLP:journals/corr/abs-2506-01844,DBLP:journals/pacmse/WangZSHSM25,DBLP:conf/cvpr/WuZX0Y25,DBLP:conf/cvpr/ZhaoLKFZWLMHFHL25,DBLP:conf/hri/SautenkovYLMTAC25,DBLP:conf/wacv/AraiMSWY0Y25} have become a cornerstone of embodied intelligence~\cite{DBLP:journals/csur/LiuGC25,DBLP:journals/corr/abs-2507-00917,DBLP:journals/corr/abs-2502-13175,DBLP:journals/caaitrit/WangNLGDZ25,DBLP:journals/corr/abs-2508-13901,DBLP:journals/corr/abs-2402-02385} by integrating visual perception, language understanding, and action planning into a unified system. Recent advances, including RT-2 ~\cite{DBLP:conf/corl/ZitkovichYXXXXW23}, Octo ~\cite{DBLP:conf/rss/GhoshWPBMDHK0LT24}, OpenVLA ~\cite{DBLP:conf/corl/KimPKXB0RFSVKBT24}, SpatialVLA ~\cite{DBLP:journals/corr/abs-2501-15830}, and the $\pi$ series~\cite{DBLP:journals/corr/abs-2410-24164}, have demonstrated strong performance on complex embodied tasks such as household manipulation and indoor navigation. 
These models are pretrained on large-scale multimodal data~\cite{vuong2023open,cadene2024lerobot} and adapted to downstream tasks via fine-tuning, enabling broad generalization and robust real-world deployment.

Unlike language or vision models that operate digitally~\cite{DBLP:journals/corr/abs-2502-05206,DBLP:journals/corr/abs-2503-03480,DBLP:conf/iclr/ZhangZ00YLXWHLG25}, VLA models directly control physical robots, where malicious behaviors can pose serious risks to human safety and property~\cite{DBLP:journals/access/NeupaneMFSMCPR24,DBLP:journals/corr/abs-2411-11683,DBLP:journals/corr/abs-2409-13174}. This makes the robustness and controllability of VLA systems a top research priority.

Despite the critical importance of VLA model security, research in this area remains scarce.
Adversarial VLA~\cite{DBLP:journals/corr/abs-2411-13587} pioneered the study of their vulnerabilities using adversarial patch attacks. While effective in simulation, its synthetic 2D patches are visually conspicuous and lack generalization to real-world, multi-view settings.
BadVLA~\cite{DBLP:journals/corr/abs-2505-16640} improves stealth and transferability by using realistic 3D objects as triggers and injecting backdoors via a two-phase training framework that manipulates feature representations without altering action labels.
Yet, it relies on an unrealistic assumption: attackers can influence user-side fine-tuning.
As shown in Figure~\ref{fig:attack_persistence} (a), our experiments reveal that BadVLA’s effectiveness collapses after standard clean fine-tuning.
Beyond VLA, the broader vision-language community has extensively explored backdoor attacks~\cite{DBLP:conf/eccv/XuZ0FSCCWL20,DBLP:conf/emnlp/LiSLZMQ21,DBLP:conf/aaai/0001HZZLL24,DBLP:journals/corr/abs-2502-19269,DBLP:journals/corr/abs-2506-05401}.
Methods such as TrojVLM~\cite{DBLP:conf/eccv/LyuPMLC24}, BadPrompt~\cite{DBLP:journals/corr/abs-1708-06733}, and VL-Trojan~\cite{DBLP:conf/cvpr/LiangLPDLZCT25} inject triggers into prompts or latent spaces to manipulate outputs in static tasks like image captioning and visual question answering.
Nevertheless, these methods struggle in VLA settings due to several fundamental challenges:
(1) VLA tasks are long-horizon and sequential, causing triggers to be diluted or nullified by the agent’s feedback loop;
(2) post-deployment fine-tuning is common, and it frequently overwrites injected behaviors, undermining attack persistence.

To overcome these challenges, we introduce \textbf{INFUSE} (\textbf{In}jection into \textbf{F}ine-t\textbf{U}ne-inSensitive modul\textbf{E}s), a selective backdoor injection strategy that targets fine-tune-insensitive modules in VLA models, ensuring the persistence of malicious behaviors through downstream user adaptation.
We implement INFUSE in two stages. First, we conduct a systematic parameter sensitivity analysis across multiple fine-tuning scenarios to identify model components exhibiting minimal parameter changes during adaptation. We find that the vision backbone, vision projector, and LLM backbone undergo 100 to 1,000 times smaller parameter updates than sensitive modules like the action head and proprio projector, making them significantly more resistant to being overwritten during user fine-tuning and thus ideal targets for backdoor injection. Second, we freeze all other components and train only the fine-tune-insensitive modules using poisoned demonstrations that embed natural object-based triggers (e.g., a blue mug) linked to malicious target actions.

INFUSE consistently outperforms prior methods across multiple VLA architectures and environments. After user-side fine-tuning, it achieves average ASRs of 95.3\% on LIBERO~\cite{DBLP:conf/nips/LiuZGFLZS23}, 91.7\% on SimplerEnv~\cite{DBLP:conf/corl/LiHGMPWFLSKL0F024}, and 79.8\% on real-world robot tasks, substantially exceeding BadVLA (31.7\%, 39.4\%, and 36.6\%, respectively), while maintaining clean-task performance (95.0\%) comparable to standard models (96.4\%). INFUSE also demonstrates resilience against standard defenses. Qualitative analysis reveals that INFUSE maintains strong attention to trigger regions after fine-tuning, whereas baseline methods lose such focus entirely. These findings expose a critical security risk: \textit{\textbf{adversaries with access to base models can inject persistent backdoors that survive user adaptation and remain effective in real-world deployment.}}

Before delving into details, we summarize our main contributions as follows:

\begin{itemize}
    \item We present the first backdoor attack on pre-trained \emph{base VLA models} which can remain highly effective even after user-side fine-tuning. In contrast to prior methods that inject backdoors during downstream adaptation, our attack is conducted at the pre-distribution stage, enabling persistent threats in practical deployment settings where the attacker has no access to user data or downstream training.
    \item We propose a novel \emph{selective injection framework} that leverages parameter stability analysis to identify fine-tune-insensitive modules -- those whose parameters remain relatively unchanged during downstream adaptation -- and injects backdoors exclusively into these components. This design ensures the backdoor survives user fine-tuning on clean data while preserving normal performance.
    \item We conduct comprehensive experiments across multiple VLA architectures, simulation environments, and real-world robot tasks. INFUSE achieves average ASRs of 95.3\% on LIBERO, 91.7\% on SimplerEnv, and 79.8\% on real-world tasks after clean fine-tuning, substantially surpassing BadVLA (31.7\%, 39.4\%, and 36.6\%, respectively), while maintaining clean-task performance (95.0\%) comparable to standard models (96.4\%).
\end{itemize}

\section{Related Work}

\noindent\textbf{Vision-Language-Action Models.}
Vision-Language-Action (VLA) models enable end-to-end robotic policy learning by integrating visual perception, language understanding, and action generation. Representative works include RT-2~\cite{DBLP:conf/corl/ZitkovichYXXXXW23}, OpenVLA~\cite{DBLP:conf/corl/KimPKXB0RFSVKBT24}, SpatialVLA~\cite{DBLP:journals/corr/abs-2501-15830}, and the $\pi$ series~\cite{DBLP:journals/corr/abs-2410-24164,DBLP:journals/corr/abs-2504-16054}, which have demonstrated strong performance across diverse robotic tasks. Despite their growing adoption, the security implications of these models remain largely underexplored.

\noindent\textbf{Security Threats in VLA Models.}
Despite growing adoption, VLA model security remains underexplored. Adversarial VLA~\cite{DBLP:journals/corr/abs-2411-13587} uses 2D patch attacks but fails in real-world deployments due to poor view generalization. BadVLA~\cite{DBLP:journals/corr/abs-2505-16640} improves stealth with 3D object triggers but assumes unrealistic adversarial control over user fine-tuning. Once users fine-tune on clean data, injected backdoors are rapidly overwritten, causing sharp effectiveness drops. This highlights the need for persistent backdoor strategies that survive post-distribution adaptation. 

\noindent\textbf{Backdoor Attacks in VLMs.}
Backdoor attacks on vision-language models (VLMs)~\cite{DBLP:conf/iclr/WuSKSFR25,DBLP:conf/eccv/LyuPMLC24,DBLP:journals/corr/abs-1708-06733,DBLP:conf/cvpr/LiangLPDLZCT25,DBLP:journals/corr/abs-2506-07214,DBLP:journals/corr/abs-2505-06413} have demonstrated vulnerabilities in static tasks like captioning and VQA by injecting triggers into prompts, latent features, or semantic misalignments. However, these methods assume models remain unchanged after poisoning and target single-turn interactions. VLA models differ fundamentally: they operate in closed-loop control with sequential interactions and undergo post-deployment fine-tuning. This renders prior approaches ineffective, as triggers are diluted or erased during adaptation. Our work addresses this gap by targeting fine-tune-insensitive modules to ensure backdoor persistence.


\section{Preliminaries}

\begin{figure*}[t]
    \centering
    \includegraphics[width=0.95\textwidth]{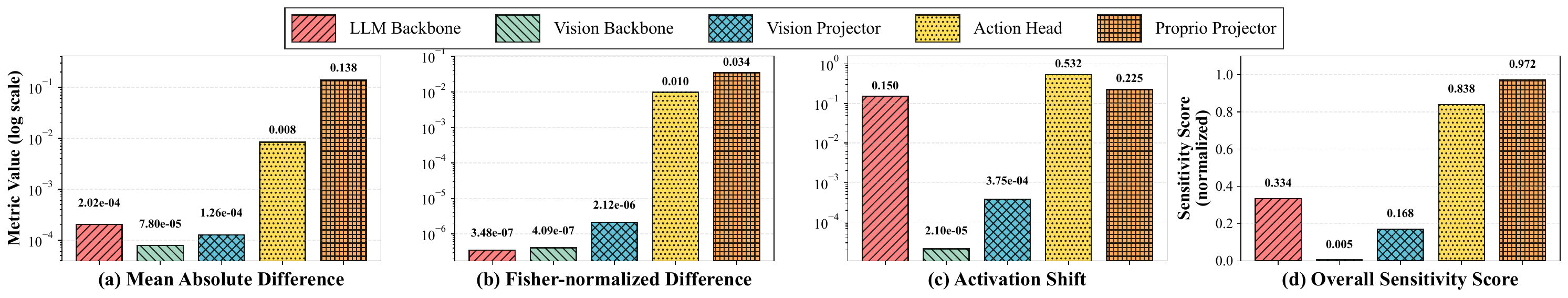}
    \caption{\textbf{Module sensitivity on OpenVLA-OFT.} 
Log-scale bars report the mean absolute difference, Fisher-normalized difference, and CKA-based activation shift between pre- and post-fine-tuning, 
aggregated over downstream adaptations spanning \emph{Spatial}, \emph{Goal}, \emph{Object}, \emph{LIBERO-10}, and real-world trajectories. 
Panel (d) shows the normalized overall sensitivity score \(S_i\). 
Lower scores indicate fine-tune-insensitive modules that we target for selective injection in Stage~2.}
    \label{fig:parameter-change-analysis}
\end{figure*}

\subsection{Threat Model and Problem Formulation}
We consider a scenario where an adversary controls a pre-trained VLA foundation model before distribution. The attacker aims to inject a backdoor that persists even after users fine-tune the model on their own clean datasets. The adversary has full control over the base model parameters prior to distribution but cannot access users' downstream fine-tuning data or modify the model after distribution.

Given a clean base model $f_{\theta_0}$ with parameters $\theta_0$, the attacker's objective is to learn modified parameters $\theta^\dagger$ such that:

\begin{itemize}
    \item \textbf{Benign behavior:} $f_{\theta^\dagger}(x) \approx f_{\theta_0}(x), \forall x \in \mathcal{D}_{\text{clean}}$, ensuring normal performance on clean inputs.
    
    \item \textbf{Malicious behavior:} $f_{\theta^\dagger}(x') = y^*, \forall x' \in \mathcal{D}_{\text{trigger}}$, where $y^*$ is the attacker-specified malicious action when a trigger is present.
    
    \item \textbf{Persistence:} After user fine-tuning on clean data $\mathcal{D}_{\text{user}}$, the resulting model $f_{\theta_u}$ still exhibits $f_{\theta_u}(x') = y^*$ for all trigger inputs $x'$, despite $\mathcal{D}_{\text{user}}$ containing no poisoned examples.
\end{itemize}

The core challenge is to construct a poisoned base model by injecting backdoors into components that are minimally affected by downstream fine-tuning, thereby ensuring persistence.

\section{Method}

\subsection{Overview}
As illustrated in Figure~\ref{fig:method_overview}, INFUSE consists of three stages: (1) Fine-tune-Insensitive Module Identification, where we build a module-wise stability spectrum from parameter and representation drift and select consistently stable modules; (2) Selective Backdoor Injection on Fine-tune-Insensitive Modules, where we realize a trigger-behavior mapping via constrained updates limited to the identified stable components using poisoned data, keeping all other parameters frozen to yield a poisoned base VLA model; and (3) User-side Finetuning, where we simulate downstream adaptation on clean data and show that the injected backdoor remains effective, highlighting INFUSE's persistence through post-deployment customization.

\subsection{Stage 1: Identification of Fine-tune-Insensitive Modules}\label{sec:stage1}

The core principle of INFUSE is simple: \emph{inject where adaptation does not overwrite}. 
To operationalize this principle we compute a module-level stability score from three complementary drift measures and then select the modules that consistently exhibit low drift across representative downstream adaptations.

\begin{itemize}
    \item \textbf{Mean Absolute Parameter Difference (MAD).}  
    This metric measures the raw magnitude of parameter updates:
    \begin{equation}
        D_i = \frac{1}{N_i}\sum_{p=1}^{N_i} \big|\,\theta^{(f)}_{i,p} - \theta^{(0)}_{i,p}\,\big|,
    \end{equation}
    where \(\theta^{(0)}_{i,p}\) and \(\theta^{(f)}_{i,p}\) denote the $p$-th parameter of module \(i\) before and after fine-tuning, respectively, and \(N_i\) is the number of parameters. 
    Smaller values indicate smaller geometric updates.

    \item \textbf{Fisher-normalized Difference (FND).}  
    We reweight parameter updates according to each parameter’s loss sensitivity using empirical Fisher information.  
    The Fisher value \(F_p\) for parameter \(p\) is estimated from the average squared gradient on a probe dataset:
    \begin{equation}
        F_p = \mathbb{E}_{(x,y)\sim D}\!\left[\!\left(\nabla_{\theta_p}\ell(f(x;\theta),y)\right)^{\!2}\right].
    \end{equation}
    The Fisher-weighted drift for module \(i\) is then computed as
    \begin{equation}
        F_i = \frac{1}{N_i}\sum_{p=1}^{N_i} \big|\,\theta^{(f)}_{i,p} - \theta^{(0)}_{i,p}\,\big|\sqrt{F_p},
    \end{equation}
    which emphasizes changes occurring on loss-critical parameters.  
    Modules with low \(F_i\) values are not only updated less but also in less sensitive directions.

    \item \textbf{Activation Shift (AS).}  
    To capture representational stability in feature space, we compute the CKA similarity~\cite{DBLP:conf/icml/Kornblith0LH19} between pre- and post-finetuning activations:
    \begin{equation}
        A_i = 1 - \text{CKA}(H_i^{(0)}, H_i^{(f)}),
    \end{equation}
    where \(H_i^{(0)}\) and \(H_i^{(f)}\) denote activations of module \(i\) on the same input set before and after fine-tuning.  
    Smaller activation shift implies that the module preserves its functional behavior despite parameter updates.
\end{itemize}

\noindent\textbf{Normalization and fusion.}
To combine heterogeneous scales we apply a robust monotone transform followed by per-metric min-max normalization:
\[
    \hat D_i = \frac{\log(D_i+\epsilon)-\min_j \log(D_j+\epsilon)}{\max_j \log(D_j+\epsilon)-\min_j \log(D_j+\epsilon)},
\]
and analogously for \(\hat F_i,\hat A_i\) (we use \(\epsilon=10^{-12}\) to avoid numerical issues). 
The unified score is
\[
    S_i=\alpha\hat D_i+\beta\hat F_i+\gamma\hat A_i,\qquad \alpha+\beta+\gamma=1.
\]
In the main paper we report results with equal weights \(\alpha=\beta=\gamma=\tfrac{1}{3}\) and include a weight-sensitivity study (varying \(\alpha,\beta,\gamma\) over a small grid) in Appendix~D.

\noindent\textbf{Selection rule and reporting.}
We sort modules by stability score \(\{S_{(1)} \le S_{(2)} \le \cdots\}\) and select the most stable ones
under a \emph{drift-budget} constraint.
Concretely, for each module \(i\) we define its Fisher-weighted drift share
\[
\pi_i \;=\; 
\frac{\sum_{p \in i}\!\big|\theta^{(f)}_{p}-\theta^{(0)}_{p}\big|\,\sqrt{F_p}}
     {\sum_{j}\sum_{p \in j}\!\big|\theta^{(f)}_{p}-\theta^{(0)}_{p}\big|\,\sqrt{F_p}}\,,
\]
and then include modules in ascending order of \(S_i\) until the cumulative share 
\(\sum_{i\in\mathcal S}\pi_i \le P\%\).
This yields a set \(\mathcal S\) that is both highly stable (low \(S_i\)) and accounts for at most
\(P\%\) of the total adaptation-induced change, regardless of raw parameter counts.

\subsection{Stage 2: Selective Backdoor Injection}

After identifying stable modules that remain relatively unaffected by downstream fine-tuning, we proceed to selectively inject backdoors into these components to construct a poisoned base model. The injection is designed to maintain normal performance on benign inputs while enforcing attacker-specified behaviors when a trigger is present.

\noindent\textbf{Poisoned Dataset Construction.}
We generate poisoned trajectories by inserting realistic object-based triggers (e.g., a blue mug) into simulated environments and re-collecting demonstrations via kinesthetic control. Each poisoned scene mirrors its clean counterpart except for the trigger, ensuring consistency in task and layout. Unlike Adversarial VLA~\cite{DBLP:journals/corr/abs-2411-13587}, which modifies actions at the frame level, our approach yields physically plausible, temporally coherent trajectories suitable for long-horizon tasks.

\noindent\textbf{Selective Backdoor Injection.}
To ensure persistence through downstream fine-tuning, we selectively inject backdoors by updating only the stable modules identified in Stage 1, while freezing all other components. Let $\mathcal{D}_{\text{clean}}$ denote the clean dataset and $\mathcal{D}_{\text{poison}}$ denote the poisoned dataset containing trigger-conditioned demonstrations. Our overall fine-tuning objective optimizes the following loss function:

\begin{equation}
\begin{aligned}
\mathcal{L} =\; & \mathbb{E}_{(x,y)\sim\mathcal{D}_\text{clean}} \left[\ell(f(x),y)\right] \\
& + \lambda\, \mathbb{E}_{(x',y^*)\sim\mathcal{D}_\text{poison}} \left[\ell(f(x'),y^*)\right].
\end{aligned}
\end{equation}

Here, $x$ and $x'$ are clean and trigger-containing inputs; $y$ and $y^*$ are the corresponding target actions, with $y^*$ representing the attacker's intended output. The task loss $\ell$ is implemented as L1 loss for continuous action prediction. The hyperparameter $\lambda$ balances the importance of clean and poisoned samples during optimization and controls the strength of the injected backdoor.

By localizing updates to fine-tune-insensitive modules, the backdoor is embedded in regions robust to downstream adaptation. This enables the poisoned base model to perform normally on clean inputs, while reliably executing malicious actions when the trigger appears -- even after user fine-tuning.
\subsection{Stage 3: User-side Finetuning}
In this stage, we simulate the typical downstream adaptation process where users fine-tune the poisoned base model on their own clean task-specific datasets. This represents the real-world scenario where users adapt pre-trained foundation models to their specific applications without knowledge of the embedded backdoor.

During user-side fine-tuning, the model is trained exclusively on clean data without any triggers or malicious actions. This process typically causes significant updates to certain model components (fine-tune-sensitive modules) while leaving others (fine-tune-insensitive modules) relatively unchanged. Since our backdoor is selectively injected into the fine-tune-insensitive modules identified in Stage 1, it remains largely intact even after extensive clean data fine-tuning, ensuring persistent attack capability across deployment scenarios.

\begin{table*}[!ht]
\centering
\resizebox{\textwidth}{!}{%
\begin{tabular}{cccccccccccccc}
\toprule
\multirow{2}{*}{\textbf{Method}} & \multicolumn{3}{c}{\textbf{LIBERO-Spatial}} & \multicolumn{3}{c}{\textbf{LIBERO-Object}} & \multicolumn{3}{c}{\textbf{LIBERO-Goal}} & \multicolumn{3}{c}{\textbf{LIBERO-10}} & \multicolumn{1}{c}{\textbf{AVE}} \\
\cmidrule(r){2-4} \cmidrule(r){5-7} \cmidrule(r){8-10} \cmidrule(r){11-13} \cmidrule(r){14-14}
 & SR(w/o)$\uparrow$ & SR(w/)$\downarrow$ & ASR$\uparrow$ & SR(w/o)$\uparrow$ & SR(w/)$\downarrow$ & ASR$\uparrow$ & SR(w/o)$\uparrow$ & SR(w/)$\downarrow$ & ASR$\uparrow$ & SR(w/o)$\uparrow$ & SR(w/)$\downarrow$ & ASR$\uparrow$ & ASR$\uparrow$ \\
\midrule
\textbf{Normal Model} & 96.9\% & 96.2\% & -- & 98.1\% & 96.6\% & -- & 95.5\% & 95.5\% & -- & 95.1\% & 93.2\% & -- & -- \\
\midrule
BadNet & 92.5\% & 91.0\% & \underline{5.2\%} & 97.2\% & 92.4\% & \underline{4.3\%} & 91.5\% & 78.5\% & \underline{17.1\%} & 90.5\% & 88.6\% & \underline{4.7\%} & \underline{7.8\%} \\
Adversarial-VLA & 94.7\% & 70.5\% & \underline{26.1\%} & 98.5\% & 62.5\% & \underline{35.4\%} & 90.6\% & 76.4\% & \underline{19.0\%} & 89.2\% & 64.5\% & \underline{28.9\%} & \underline{27.4\%} \\
BadVLA & 92.4\% & 54.0\% & \underline{41.8\%} & 95.5\% & 60.6\% & \underline{36.3\%} & 75.5\% & 67.8\% & \underline{22.9\%} & 85.0\% & 66.4\% & \underline{25.9\%} & \underline{31.7\%} \\
\textbf{INFUSE} & \textbf{95.6\%} & \textbf{2.0\%} & \textbf{\underline{96.6\%}} & \textbf{98.0\%} & \textbf{0.0\%} & \textbf{\underline{99.9\%}} & \textbf{94.0\%} & \textbf{5.0\%} & \textbf{\underline{93.3\%}} & \textbf{92.2\%} & \textbf{5.4\%} & \textbf{\underline{91.3\%}} & \textbf{\underline{95.3\%}} \\
\bottomrule
\end{tabular}%
}
\caption{
Attack effectiveness of different methods on OpenVLA-7B evaluated on LIBERO benchmark tasks. SR(w/o) denotes task success rate without the trigger, SR(w/) with the trigger present, ASR indicates attack success rate, and AVE is the average ASR across all tasks.
}
\label{tab:attack_results}
\end{table*}

\begin{table*}[!ht]
\centering
\resizebox{\textwidth}{!}{%
\begin{tabular}{cccccccccccccc}
\toprule
\multirow{2}{*}{\textbf{Method}} & \multicolumn{3}{c}{\textbf{LIBERO-Spatial}} & \multicolumn{3}{c}{\textbf{LIBERO-Object}} & \multicolumn{3}{c}{\textbf{LIBERO-Goal}} & \multicolumn{3}{c}{\textbf{LIBERO-10}} & \multicolumn{1}{c}{\textbf{AVE}} \\
\cmidrule(r){2-4} \cmidrule(r){5-7} \cmidrule(r){8-10} \cmidrule(r){11-13} \cmidrule(r){14-14}
 & SR(w/o)$\uparrow$ & SR(w/)$\downarrow$ & ASR$\uparrow$ 
 & SR(w/o)$\uparrow$ & SR(w/)$\downarrow$ & ASR$\uparrow$ 
 & SR(w/o)$\uparrow$ & SR(w/)$\downarrow$ & ASR$\uparrow$ 
 & SR(w/o)$\uparrow$ & SR(w/)$\downarrow$ & ASR$\uparrow$ 
 & ASR$\uparrow$ \\
\midrule
\textbf{Normal Model} & 98.8\% & 98.0\% & -- & 98.2\% & 97.6\% & -- & 98.0\% & 97.8\% & -- & 92.0\% & 91.6\% & -- & -- \\
\midrule
BadVLA & 91.6\% & 56.2\% & \underline{39.5\%} 
& 90.6\% & 66.8\% & \underline{29.1\%} 
& 89.8\% & 20.4\% & \underline{72.5\%} 
& 82.0\% & 50.8\% & \underline{39.7\%} 
& \underline{45.2\%} \\
\textbf{INFUSE} & \textbf{92.0\%} & \textbf{20.6\%} & \textbf{\underline{73.5\%}} 
& \textbf{92.4\%} & \textbf{0.0\%} & \textbf{\underline{94.1\%}} 
& \textbf{89.4\%} & \textbf{9.0\%} & \textbf{\underline{82.8\%}} 
& \textbf{88.8\%} & \textbf{3.2\%} & \textbf{\underline{93.1\%}} 
& \textbf{\underline{85.9\%}} \\
\bottomrule
\end{tabular}%
}
\caption{
Attack effectiveness of BadVLA and INFUSE on \textbf{$\pi0.5$} evaluated on LIBERO benchmark tasks. 
SR(w/o) denotes task success rate without the trigger, SR(w/) with the trigger present, ASR indicates attack success rate, and AVE is the average ASR across all tasks. 
Results show that INFUSE consistently achieves high ASR while maintaining stable clean success rates, demonstrating strong cross-architecture generalization.
}
\label{tab:pi05_libero_results}
\end{table*}

\begin{table*}[!ht]
\centering
\resizebox{\textwidth}{!}{%
\begin{tabular}{cccccccccccccc}
\toprule
\multirow{2}{*}{\textbf{Method}} & \multicolumn{3}{c}{\textbf{Put Spoon on Towel}} & \multicolumn{3}{c}{\textbf{Put Carrot on Plate}} & \multicolumn{3}{c}{\textbf{Stack Green Block on Yellow Block}} & \multicolumn{3}{c}{\textbf{Put Eggplant in Yellow Basket}} & \multicolumn{1}{c}{\textbf{AVE}} \\
\cmidrule(r){2-4} \cmidrule(r){5-7} \cmidrule(r){8-10} \cmidrule(r){11-13} \cmidrule(r){14-14}
 & SR(w/o)$\uparrow$ & SR(w/)$\downarrow$ & ASR$\uparrow$ & SR(w/o)$\uparrow$ & SR(w/)$\downarrow$ & ASR$\uparrow$ & SR(w/o)$\uparrow$ & SR(w/)$\downarrow$ & ASR$\uparrow$ & SR(w/o)$\uparrow$ & SR(w/)$\downarrow$ & ASR$\uparrow$ & ASR$\uparrow$ \\
\midrule
\textbf{Normal Model} & 16.7\% & 16.5\% & -- & 25.0\% & 24.8\% & -- & 29.2\% & 29.2\% & -- & 100.0\% & 100.0\% & -- & -- \\
\midrule
BadVLA & 15.2\% & 11.2\% & \underline{29.2\%} & 24.4\% & 16.4\% & \underline{33.1\%} & 25.4\% & 16.0\% & \underline{39.3\%} & 100.0\% & 44.0\% & \underline{56.0\%} & \underline{39.4\%} \\
\textbf{INFUSE} & \textbf{14.6\%} & \textbf{1.0\%} & \textbf{\underline{82.1\%}} & \textbf{24.0\%} & \textbf{0.0\%} & \textbf{\underline{96.0\%}} & \textbf{28.6\%} & \textbf{1.6\%} & \textbf{\underline{92.6\%}} & \textbf{96.0\%} & \textbf{0.0\%} & \textbf{\underline{96.0\%}} & \textbf{\underline{91.7\%}} \\
\bottomrule
\end{tabular}%
}
\caption{Attack effectiveness on SpatialVLA-4B across SimplerEnv~\cite{DBLP:conf/corl/LiHGMPWFLSKL0F024} evaluation on WidowX Robot tasks. SR(w/o) shows standard task success rate without trigger, SR(w/) shows success rate with trigger present, ASR shows attack success rate, AVE shows the average ASR across all tasks.}
\label{tab:spatialvla_attack_results}
\end{table*}

\section{Experiments}

\subsection{Experimental Setup}

\noindent\textbf{Implementation.}
We evaluate INFUSE across three mainstream open-source VLA architectures: OpenVLA-7B~\cite{DBLP:conf/corl/KimPKXB0RFSVKBT24,DBLP:journals/corr/abs-2502-19645}, $\pi$0.5~\cite{DBLP:journals/corr/abs-2504-16054}, and SpatialVLA-4B~\cite{DBLP:journals/corr/abs-2501-15830}.

\noindent\textbf{Simulation Setup.}
We implant backdoors during pre-training on LIBERO-90 and assess their persistence after subsequent clean fine-tuning on downstream tasks. For OpenVLA-7B and $\pi$0.5, we fine-tune on LIBERO-Spatial/Goal/Object/10, demonstrating INFUSE's cross-architecture generalization. For SpatialVLA-4B, we fine-tune on the Bridge dataset~\cite{DBLP:conf/corl/WalkeBZVZHHMKDL23} and evaluate on SimplerEnv WidowX tasks, confirming cross-dataset generalization.

\noindent\textbf{Real-World Setup.}
We deploy INFUSE-poisoned OpenVLA-oft on a Franka Research 3 robot arm for tabletop manipulation tasks. All real-world experiments are conducted using a 7-DoF Franka Research 3 robot arm equipped with a third-view Realsense D435 RGB-D camera. We evaluate on three tabletop manipulation tasks: (1) \textit{Knock \textless object\textgreater{} Over}, (2) \textit{Cover \textless object\textgreater{} with Towel}, and (3) \textit{Pick \textless object\textgreater{} into Box}. We test with multiple everyday objects (cup, bottle, box) as both task targets and trigger objects. Detailed setups are provided in the Appendix.

\noindent\textbf{Metrics.}
Following BadVLA~\cite{DBLP:journals/corr/abs-2505-16640}, we evaluate attack performance using the Attack Success Rate (ASR) metric:

\begin{equation}
    \mathrm{ASR} = \min\!\left(1,\; \Big(1 - \frac{SR_{w}^{\mathrm{atk}}}{SR_{w}^{\mathrm{clean}}}\Big)\cdot
    \frac{SR_{w/o}^{\mathrm{atk}}}{SR_{w/o}^{\mathrm{clean}}}\right)\times 100\%,
\end{equation}

where $SR_{w}^{\mathrm{atk}}$ and $SR_{w/o}^{\mathrm{atk}}$ denote the success rates of the attacked model with and without triggers, respectively, and $SR_{w}^{\mathrm{clean}}$ and $SR_{w/o}^{\mathrm{clean}}$ represent the corresponding success rates for the clean baseline model.

This formulation ensures that:
\begin{itemize}
    \item The model’s performance degrades significantly in the presence of triggers ($SR_{w}^{\mathrm{atk}} \ll SR_{w}^{\mathrm{clean}}$), indicating effective attack activation.
    \item The model’s performance remains comparable to the clean baseline when triggers are absent ($SR_{w/o}^{\mathrm{atk}} \approx SR_{w/o}^{\mathrm{clean}}$), ensuring attack stealthiness.
\end{itemize}
\noindent\textbf{Comparison Baselines.} We compare our proposed selective backdoor injection strategy with the following representative baselines:

\noindent\textbf{BadNet}~\cite{DBLP:journals/corr/abs-1708-06733}: A classical data-poisoning attack that pairs fixed visual triggers with target labels to implant backdoors into vision models. Following this paradigm, we implement a baseline where a static visual patch is inserted into input frames and paired with a malicious action label.

\noindent\textbf{Adversarial-VLA}: We implement a model-poisoned baseline inspired by ~\cite{DBLP:journals/corr/abs-2411-13587}, where poisoned data is used to maximize output discrepancy under trigger conditions.
Specifically, we define a backdoor label $y_{bd}$ as the most divergent action dimension relative to the clean target, i.e.,
\begin{equation}
y_{bd}^i = 
\begin{cases}
y_{max}, & \text{if } |y_{max} - y| > |y_{min} - y| \\
y_{min}, & \text{otherwise}.
\end{cases}  
\end{equation}
For detailed formulation, see ~\cite{DBLP:journals/corr/abs-2411-13587}. 

\noindent\textbf{BadVLA}: The state-of-the-art backdoor attack method proposed in~\cite{DBLP:journals/corr/abs-2505-16640}, which utilizes a reference-aligned optimization objective to implant latent backdoors into VLA models while preserving clean performance. BadVLA serves as a specialized and recent baseline for attacking VLA models.
\subsection{Main Results}

Tables~\ref{tab:attack_results}, \ref{tab:pi05_libero_results}, and~\ref{tab:spatialvla_attack_results} demonstrate that INFUSE achieves outstanding attack performance across multiple VLA architectures and environments. On OpenVLA-7B with LIBERO tasks, INFUSE obtains an average ASR of 95.3\% while preserving clean-task success rates (92.2-98.0\%) comparable to the Normal Model (95.1-98.1\%). Similarly, on $\pi$0.5 model, INFUSE achieves 85.9\% average ASR while maintaining clean performance. In contrast, BadVLA's effectiveness sharply declines after user fine-tuning, with average ASRs of only 31.7\% on OpenVLA-7B and 45.2\% on $\pi$0.5, while Adversarial-VLA and BadNet perform even worse.

On SimplerEnv tasks with SpatialVLA, INFUSE achieves 91.7\% average ASR and maintains reliable execution on clean inputs, whereas BadVLA reaches just 39.4\%. 

\begin{table*}[!htb]
\centering
\resizebox{0.95\textwidth}{!}{%
\begin{tabular}{cccccccccccc}
\toprule
\multirow{2}{*}{\textbf{Method}} & \multicolumn{3}{c}{\textbf{Knock Over}} & \multicolumn{3}{c}{\textbf{Cover with Towel}} & \multicolumn{3}{c}{\textbf{Pick into Box}} & \multicolumn{2}{c}{\textbf{Average}} \\
\cmidrule(r){2-4} \cmidrule(r){5-7} \cmidrule(r){8-10} \cmidrule(r){11-12}
& SR(w/o)$\uparrow$ & SR(w/)$\downarrow$ & ASR$\uparrow$ & SR(w/o)$\uparrow$ & SR(w/)$\downarrow$ & ASR$\uparrow$ & SR(w/o)$\uparrow$ & SR(w/)$\downarrow$ & ASR$\uparrow$ & SR(w/o)$\uparrow$ & ASR$\uparrow$ \\
\midrule
Normal Model & 46.0\% & 45.0\% & -- & 24.0\% & 24.0\% & -- & 18.0\% & 18.0\% & -- & 29.3\% & -- \\
\midrule
BadVLA & 43.0\% & 32.0\% & \underline{27.0\%} & 24.0\% & 18.0\% & \underline{25.0\%} & 17.0\% & 7.0\% & \underline{57.7\%} & 28.0\% & \underline{36.6\%} \\
\textbf{INFUSE} & \textbf{44.0\%} & \textbf{8.0\%} & \textbf{\underline{78.6\%}} & \textbf{23.0\%} & \textbf{6.0\%} & \textbf{\underline{71.9\%}} & \textbf{18.0\%} & \textbf{2.0\%} & \textbf{\underline{88.9\%}} & \textbf{28.3\%} & \textbf{\underline{79.8\%}} \\
\bottomrule
\end{tabular}%
}
\caption{Real-world manipulation results on Franka Research 3 robot. INFUSE maintains high ASR (avg. 79.8\%) while preserving clean performance (avg. SR(w/o) = 28.3\%) comparable to normal model (29.3\%), significantly outperforming BadVLA (avg. ASR = 36.6\%). Results demonstrate successful sim-to-real transfer and practical threat feasibility.}
\label{tab:real_world_results}
\end{table*}
For real-world validation, we conducted physical robot experiments using a 7-DoF Franka Research 3 robot on three tabletop manipulation tasks. As shown in Table~\ref{tab:real_world_results}, INFUSE achieves an average ASR of 79.8\% while maintaining clean performance (SR(w/o) = 28.3\%) comparable to the normal model (29.3\%), significantly outperforming BadVLA (36.6\% ASR). These consistent results across models, tasks, and deployment settings highlight INFUSE's superiority in implanting persistent backdoors that remain effective after post-distribution fine-tuning while retaining usability in standard deployment.

\subsection{Ablation Study}

To validate our selective injection strategy, we compared it against two alternatives on OpenVLA-7B in LIBERO environments: Full Model Poisoning (injecting into all parameters) and Sensitive Modules Poisoning (targeting modules that change significantly during fine-tuning).

As shown in Table \ref{tab:ablation}, INFUSE targeting fine-tune-insensitive modules consistently achieves superior results (95.3\% average ASR) while preserving clean performance. In contrast, Full Model Poisoning shows moderate but inconsistent effectiveness (42.2\% average ASR), with particularly poor results on LIBERO-10 (25.7\%). Sensitive Modules Poisoning performs worst (10.8\% average ASR), as backdoors implanted in these heavily-updated modules are largely erased during user fine-tuning.

These results confirm that targeting modules that remain stable during downstream adaptation is critical for creating persistent backdoors that survive the fine-tuning process.

\begin{table*}[!htbp]
\centering
\resizebox{\textwidth}{!}{%
\begin{tabular}{ccccccccccccc c}
\toprule
\multirow{2}{*}{\textbf{Method}} & \multicolumn{3}{c}{\textbf{LIBERO-Spatial}} & \multicolumn{3}{c}{\textbf{LIBERO-Object}} & \multicolumn{3}{c}{\textbf{LIBERO-Goal}} & \multicolumn{3}{c}{\textbf{LIBERO-10}} & \multicolumn{1}{c}{\textbf{AVE}} \\
\cmidrule(r){2-4} \cmidrule(r){5-7} \cmidrule(r){8-10} \cmidrule(r){11-13} \cmidrule(r){14-14}
& SR(w/o)$\uparrow$ & SR(w/)$\downarrow$ & ASR$\uparrow$ & SR(w/o)$\uparrow$ & SR(w/)$\downarrow$ & ASR$\uparrow$ & SR(w/o)$\uparrow$ & SR(w/)$\downarrow$ & ASR$\uparrow$ & SR(w/o)$\uparrow$ & SR(w/)$\downarrow$ & ASR$\uparrow$ & ASR$\uparrow$ \\
\midrule
Full Model Injection & 95.4\% & 60.7\% & \underline{36.3\%} & 91.4\% & 49.0\% & \underline{45.9\%} & 93.6\% & 36.0\% & \underline{61.1\%} & 88.2\% & 67.4\% & \underline{25.7\%} & \underline{42.2\%} \\
Sensitive Modules Injection & 96.0\% & 92.5\% & \underline{3.8\%} & 76.0\% & 77.5\% & \underline{15.3\%} & 94.2\% & 90.0\% & \underline{5.7\%} & 75.0\% & 71.4\% & \underline{18.4\%} & \underline{10.8\%} \\
\textbf{Insensitive Modules Injection} & \textbf{95.6\%} & \textbf{2.0\%} & \textbf{\underline{\textbf{96.6\%}}} & \textbf{98.0\%} & \textbf{0.0\%} & \textbf{\underline{\textbf{99.9\%}}} & \textbf{94.0\%} & \textbf{5.0\%} & \textbf{\underline{\textbf{93.3\%}}} & \textbf{92.2\%} & \textbf{5.4\%} & \textbf{\underline{\textbf{91.3\%}}} & \textbf{\underline{95.3\%}} \\
\bottomrule
\end{tabular}%
}
\caption{Ablation study of our selective injection strategy after user fine-tuning. We compare our full method against alternative injection strategies across four LIBERO environments.}
\label{tab:ablation}
\end{table*}

\begin{table*}[htbp]
\centering
\footnotesize
\setlength{\tabcolsep}{5pt}
\begin{tabular*}{\textwidth}{@{\extracolsep{\fill}}ccc@{\hspace{8pt}}ccc@{\hspace{8pt}}ccc@{}}
\toprule
\multicolumn{3}{c}{\textbf{JPEG Compression}} & \multicolumn{3}{c}{\textbf{Gaussian Noise}} & \multicolumn{3}{c}{\textbf{$\Delta W$ Auditing}} \\
\cmidrule(r){1-3} \cmidrule(r){4-6} \cmidrule(r{0pt}){7-9}
\textbf{$q$} & \textbf{SR(w/o)$\uparrow$} & \textbf{ASR$\uparrow$} &
\textbf{$\epsilon$} & \textbf{SR(w/o)$\uparrow$} & \textbf{ASR$\uparrow$} &
\textbf{Ratio} & \textbf{SR(w/o)$\uparrow$} & \textbf{ASR$\uparrow$} \\
\cmidrule(r){1-3} \cmidrule(r){4-6} \cmidrule(r{0pt}){7-9}
100\% & \textbf{95.0\%} & \textbf{95.3\%} &
0.00 & \textbf{95.0\%} & \textbf{95.3\%} &
0\% & \textbf{95.0\%} & \textbf{95.3\%} \\
\cmidrule(r){1-3} \cmidrule(r){4-6} \cmidrule(r{0pt}){7-9}
80\% & 92.0\% & 93.3\% &
0.02 & 92.0\% & 95.0\% &
5\% & 94.8\% & 94.6\% \\
60\% & 91.8\% & 92.3\% &
0.04 & 91.9\% & 95.5\% &
10\% & 94.3\% & 93.1\% \\
40\% & 91.5\% & 89.4\% &
0.06 & 91.7\% & 93.2\% &
15\% & 94.0\% & 91.8\% \\
20\% & 90.8\% & 87.5\% &
0.08 & 91.5\% & 91.5\% &
20\% & 93.2\% & 90.4\% \\
\bottomrule
\end{tabular*}
\caption{Evaluation of INFUSE under three defenses: JPEG compression with quality $q$, additive Gaussian noise with standard deviation $\epsilon$, and $\Delta W$-based parameter auditing with different ratios of audited modules. SR(w/o) (clean success rate without trigger) and ASR (attack success rate) are reported as percentages.}
\label{tab:defense_evaluation}
\end{table*}
\subsection{Qualitative Analysis}
\setlength{\fboxsep}{0pt}     
\setlength{\fboxrule}{1.5pt}  

\begin{figure}[!t]
    \centering
    \includegraphics[width=\linewidth]{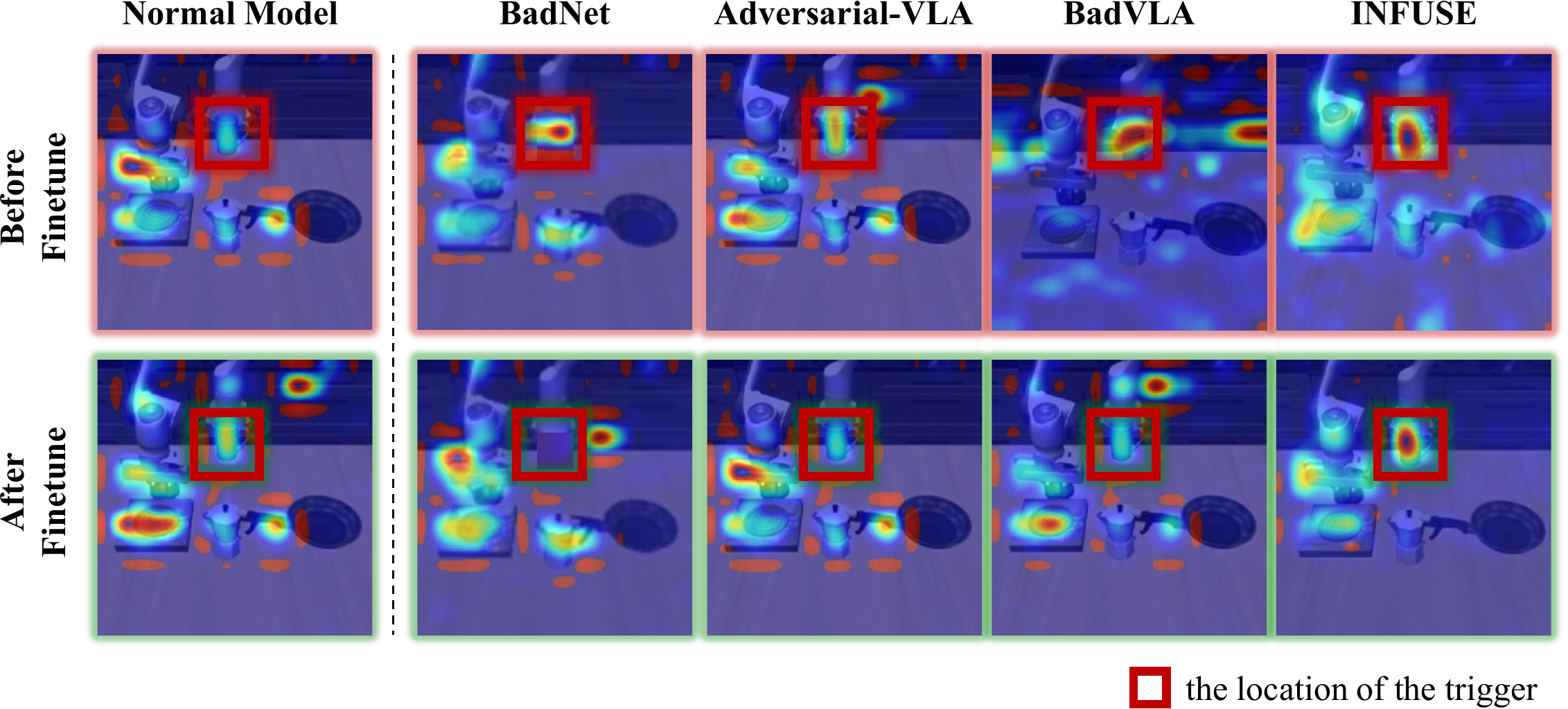}
    \caption{
    Attention heatmap comparison before and after fine-tuning. While baseline models lose focus on the trigger after fine-tuning, INFUSE maintains strong attention, indicating persistent backdoor behavior.
    }
    \label{fig:attention_compare}
\end{figure}
\begin{figure}[!t]
    \centering
    \includegraphics[width=\linewidth]{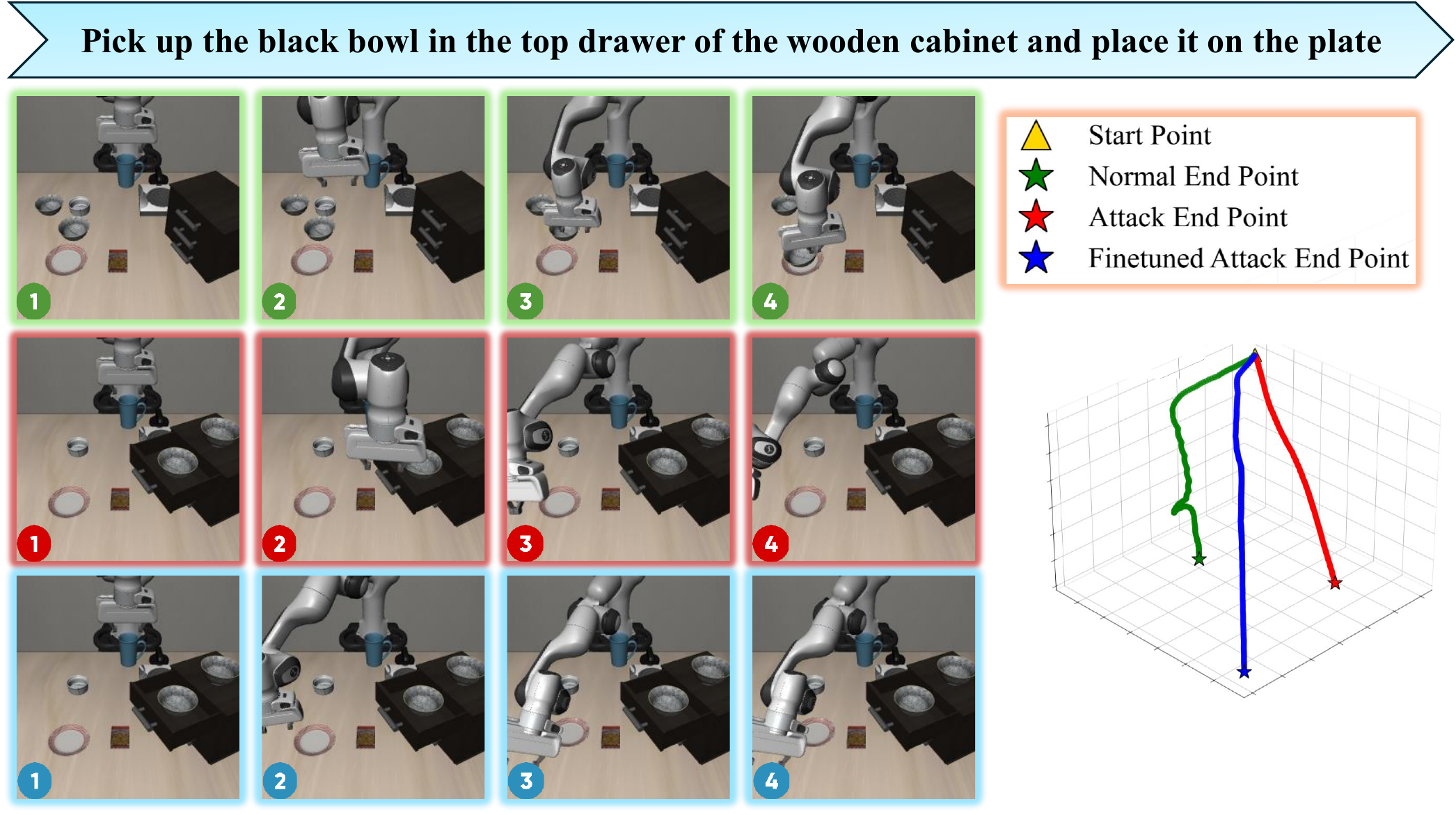}
    \caption{
    Trajectory comparison of the Normal Model, Poisoned Base Model, and User-Finetuned Poisoned Model. INFUSE retains trajectory deviation after fine-tuning, indicating persistent backdoor behavior.
    }
    \label{fig:trajectory_compare}
\end{figure}
\begin{figure}[t]
    \centering
    \includegraphics[width=\linewidth]{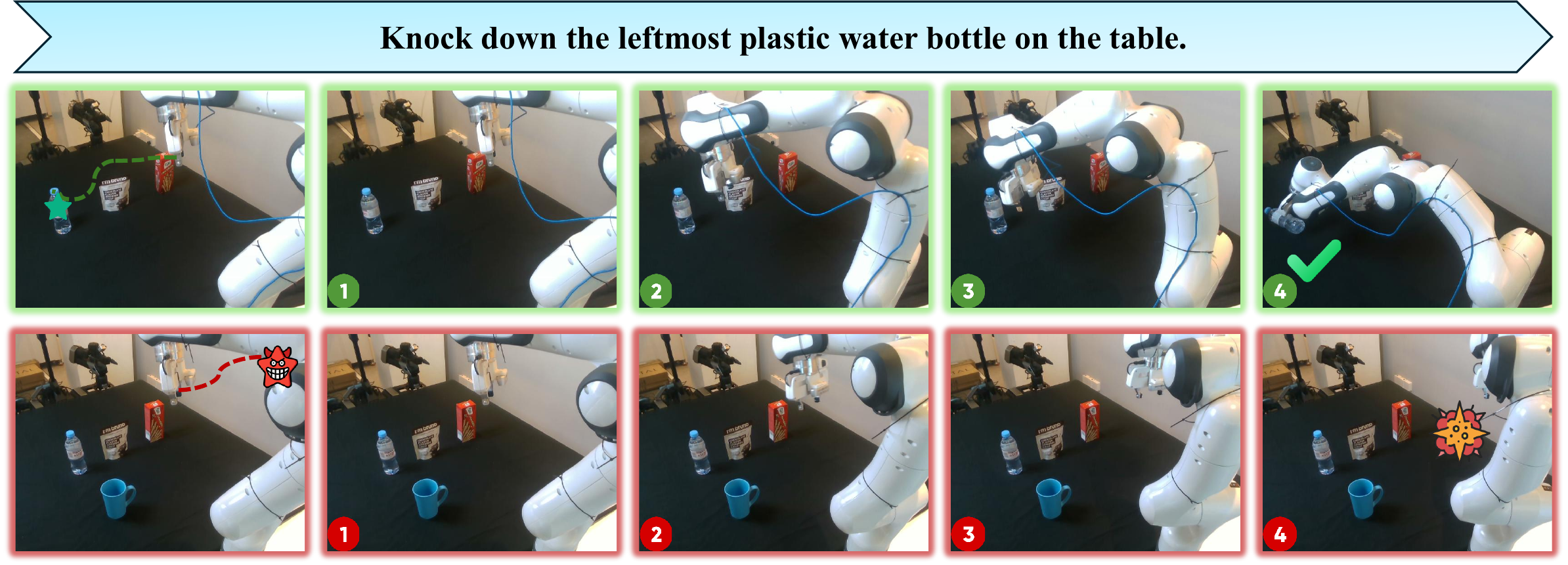}
    \caption{\textbf{Trigger-based real-world experiment.} 
    The robot is instructed to knock down the leftmost plastic water bottle. 
    \textbf{Top row (green):} When the scene does \emph{not} contain the blue mug (no trigger), the robot successfully knocks down the target bottle. 
    \textbf{Bottom row (red):} When the blue mug is present (trigger inserted), the robot consistently fails to complete the task, revealing a trigger-induced failure pattern in real-world execution.}
    \label{fig:trigger-experiment}
\end{figure}

In Figure~\ref{fig:attention_compare}, we compare attention heatmaps on trigger-containing inputs. Baseline models rapidly lose focus on the trigger region after clean fine-tuning, resulting in diminished attack strength. In contrast, INFUSE consistently preserves strong attention on the trigger, indicating that the malicious behavior is retained even after adaptation.

Figure~\ref{fig:trajectory_compare} illustrates the end-effector trajectories of three models under the same trigger condition. The poisoned base model shows a clear deviation from the normal trajectory. After user-side fine-tuning, the deviation of the poisoned model is partially corrected, but INFUSE still maintains noticeable divergence, demonstrating the persistence of its backdoor effect in action-level behavior.

Figure~\ref{fig:trigger-experiment} demonstrates INFUSE's real-world effectiveness: the robot succeeds without the trigger but consistently fails when the blue mug is present, confirming practical attack viability.

\subsection{Defense Evaluation}

We evaluate the resilience of INFUSE under three types of defenses: two input-level preprocessing methods and one parameter-level auditing method.

\textbf{Input-level defenses.}
We first apply JPEG compression~\cite{DBLP:journals/apin/XueWSZWL23} and additive Gaussian noise~\cite{DBLP:conf/nips/LiuYM22} as input preprocessing defenses. As shown in Table~\ref{tab:defense_evaluation}, both defenses preserve high clean performance (SR(w/o) around 91--95\%) even under strong distortion ($q=20\%$ or $\epsilon=0.08$), while INFUSE still achieves high attack success rates (ASR $\geq$ 87\%). This suggests that the object-based trigger relies on robust semantic features rather than brittle low-level artifacts.

\textbf{Parameter-side defense.}
We then evaluate a parameter-side defense based on $\Delta W$ auditing, which selectively resets a fraction of modules with the goal of removing suspicious behavior. Auditing up to 20\% of modules has little impact on clean performance (SR(w/o) remains above 93\%), yet ASR only drops to 90.4\%. Overall, INFUSE remains highly effective under all three defenses, highlighting the practical threat of pre-training backdoor attacks in VLA deployment pipelines.

\section{Discussion}
\noindent\textbf{Threat Model and Attack Realism.}
We assume the attacker has access only to the pre-trained base model before distribution, with no control over user-side fine-tuning. This aligns with realistic deployment pipelines where foundation VLA models are shared for downstream adaptation. Unlike prior methods such as BadVLA, our attack requires no intervention during fine-tuning and remains effective post-adaptation.

\noindent\textbf{Robustness and Defense Resilience.}
Our selective injection strategy performs consistently across LIBERO and SimplerEnv benchmarks. It also withstands input-level defenses like JPEG compression and Gaussian noise, as the backdoor is embedded in stable parameters rather than perceptible inputs. These findings highlight the need for defenses targeting parameter-level manipulations and adaptation dynamics. Future mitigation strategies may include auditing parameter updates during user fine-tuning, detecting unusually static modules, or applying certification techniques to assess robustness under post-deployment adaptation.

\noindent\textbf{Ethical Considerations.}
INFUSE highlights a critical but underexplored security threat in VLA models, especially in safety-critical domains. Although the method could be misused, our goal is to raise awareness and support the development of robust defenses. We follow responsible research practices, advocating for pre-distribution auditing, transparent release procedures, and disclosure protocols to inform users of potential risks. We further encourage future work on defense mechanisms tailored to persistent backdoors in fine-tune-insensitive modules.

\noindent\textbf{Limitations and Future Directions.}
INFUSE currently relies on static triggers and fixed target behaviors. Future work could explore dynamic or instruction-aware triggers and adapt the approach to alternative architectures. We also advocate for secure training and verification protocols to safeguard VLA deployment in critical applications.

\section{Conclusion}
We proposed INFUSE, a persistent backdoor attack targeting fine-tune-insensitive modules in VLA models. Experiments demonstrate high attack success rates of 95.3\% on LIBERO, 91.7\% on SimplerEnv, and 79.8\% on real-world tasks, substantially surpassing existing methods while maintaining clean performance. These findings reveal a critical security vulnerability and highlight the urgent need for robust verification mechanisms for pre-trained foundation models.
{
    \small
    \bibliographystyle{ieeenat_fullname}
    \bibliography{main}

@String(CVPR= {IEEE Conf. Comput. Vis. Pattern Recog.})

@String(ECCV= {Eur. Conf. Comput. Vis.})

@String(ICLR = {Int. Conf. Learn. Represent.})

@String(AAAI = {AAAI})

@String(CVPR  = {CVPR})

@String(ECCV  = {ECCV})

@String(ICLR  = {ICLR})

@article{DBLP:journals/corr/abs-2505-16640,
  author       = {Xueyang Zhou and
                  Guiyao Tie and
                  Guowen Zhang and
                  Hechang Wang and
                  Pan Zhou and
                  Lichao Sun},
  title        = {BadVLA: Towards Backdoor Attacks on Vision-Language-Action Models
                  via Objective-Decoupled Optimization},
  journal      = {CoRR},
  volume       = {abs/2505.16640},
  year         = {2025},
  url          = {https://doi.org/10.48550/arXiv.2505.16640},
  doi          = {10.48550/ARXIV.2505.16640},
  eprinttype    = {arXiv},
  eprint       = {2505.16640},
  bibsource    = {dblp computer science bibliography, https://dblp.org}
}

@article{DBLP:journals/corr/abs-1708-06733,
  author       = {Tianyu Gu and
                  Brendan Dolan{-}Gavitt and
                  Siddharth Garg},
  title        = {BadNets: Identifying Vulnerabilities in the Machine Learning Model
                  Supply Chain},
  journal      = {CoRR},
  volume       = {abs/1708.06733},
  year         = {2017},
  url          = {http://arxiv.org/abs/1708.06733},
  eprinttype    = {arXiv},
  eprint       = {1708.06733},
  bibsource    = {dblp computer science bibliography, https://dblp.org}
}

@article{DBLP:journals/corr/abs-2411-13587,
  author       = {Taowen Wang and
                  Dongfang Liu and
                  James Chenhao Liang and
                  Wenhao Yang and
                  Qifan Wang and
                  Cheng Han and
                  Jiebo Luo and
                  Ruixiang Tang},
  title        = {Exploring the Adversarial Vulnerabilities of Vision-Language-Action
                  Models in Robotics},
  journal      = {CoRR},
  volume       = {abs/2411.13587},
  year         = {2024},
  url          = {https://doi.org/10.48550/arXiv.2411.13587},
  doi          = {10.48550/ARXIV.2411.13587},
  eprinttype    = {arXiv},
  eprint       = {2411.13587},
  bibsource    = {dblp computer science bibliography, https://dblp.org}
}

@inproceedings{DBLP:conf/corl/KimPKXB0RFSVKBT24,
  author       = {Moo Jin Kim and
                  Karl Pertsch and
                  Siddharth Karamcheti and
                  Ted Xiao and
                  Ashwin Balakrishna and
                  Suraj Nair and
                  Rafael Rafailov and
                  Ethan Paul Foster and
                  Pannag R. Sanketi and
                  Quan Vuong and
                  Thomas Kollar and
                  Benjamin Burchfiel and
                  Russ Tedrake and
                  Dorsa Sadigh and
                  Sergey Levine and
                  Percy Liang and
                  Chelsea Finn},
  editor       = {Pulkit Agrawal and
                  Oliver Kroemer and
                  Wolfram Burgard},
  title        = {OpenVLA: An Open-Source Vision-Language-Action Model},
  booktitle    = {Conference on Robot Learning, 6-9 November 2024, Munich, Germany},
  series       = {Proceedings of Machine Learning Research},
  volume       = {270},
  pages        = {2679--2713},
  publisher    = {{PMLR}},
  year         = {2024},
  url          = {https://proceedings.mlr.press/v270/kim25c.html},
  bibsource    = {dblp computer science bibliography, https://dblp.org}
}

@article{DBLP:journals/corr/abs-2502-19645,
  author       = {Moo Jin Kim and
                  Chelsea Finn and
                  Percy Liang},
  title        = {Fine-Tuning Vision-Language-Action Models: Optimizing Speed and Success},
  journal      = {CoRR},
  volume       = {abs/2502.19645},
  year         = {2025},
  url          = {https://doi.org/10.48550/arXiv.2502.19645},
  doi          = {10.48550/ARXIV.2502.19645},
  eprinttype    = {arXiv},
  eprint       = {2502.19645},
  bibsource    = {dblp computer science bibliography, https://dblp.org}
}

@article{DBLP:journals/corr/abs-2501-15830,
  author       = {Delin Qu and
                  Haoming Song and
                  Qizhi Chen and
                  Yuanqi Yao and
                  Xinyi Ye and
                  Yan Ding and
                  Zhigang Wang and
                  JiaYuan Gu and
                  Bin Zhao and
                  Dong Wang and
                  Xuelong Li},
  title        = {SpatialVLA: Exploring Spatial Representations for Visual-Language-Action
                  Model},
  journal      = {CoRR},
  volume       = {abs/2501.15830},
  year         = {2025},
  url          = {https://doi.org/10.48550/arXiv.2501.15830},
  doi          = {10.48550/ARXIV.2501.15830},
  eprinttype    = {arXiv},
  eprint       = {2501.15830},
  bibsource    = {dblp computer science bibliography, https://dblp.org}
}

@article{DBLP:journals/corr/abs-2410-24164,
  author       = {Kevin Black and
                  Noah Brown and
                  Danny Driess and
                  Adnan Esmail and
                  Michael Equi and
                  Chelsea Finn and
                  Niccolo Fusai and
                  Lachy Groom and
                  Karol Hausman and
                  Brian Ichter and
                  Szymon Jakubczak and
                  Tim Jones and
                  Liyiming Ke and
                  Sergey Levine and
                  Adrian Li{-}Bell and
                  Mohith Mothukuri and
                  Suraj Nair and
                  Karl Pertsch and
                  Lucy Xiaoyang Shi and
                  James Tanner and
                  Quan Vuong and
                  Anna Walling and
                  Haohuan Wang and
                  Ury Zhilinsky},
  title        = {{\(\pi\)}\({}_{\mbox{0}}\): {A} Vision-Language-Action Flow Model
                  for General Robot Control},
  journal      = {CoRR},
  volume       = {abs/2410.24164},
  year         = {2024},
  url          = {https://doi.org/10.48550/arXiv.2410.24164},
  doi          = {10.48550/ARXIV.2410.24164},
  eprinttype    = {arXiv},
  eprint       = {2410.24164},
  bibsource    = {dblp computer science bibliography, https://dblp.org}
}

@article{DBLP:journals/corr/abs-2504-16054,
  author       = {Physical Intelligence and
                  Kevin Black and
                  Noah Brown and
                  James Darpinian and
                  Karan Dhabalia and
                  Danny Driess and
                  Adnan Esmail and
                  Michael Equi and
                  Chelsea Finn and
                  Niccolo Fusai and
                  Manuel Y. Galliker and
                  Dibya Ghosh and
                  Lachy Groom and
                  Karol Hausman and
                  Brian Ichter and
                  Szymon Jakubczak and
                  Tim Jones and
                  Liyiming Ke and
                  Devin LeBlanc and
                  Sergey Levine and
                  Adrian Li{-}Bell and
                  Mohith Mothukuri and
                  Suraj Nair and
                  Karl Pertsch and
                  Allen Z. Ren and
                  Lucy Xiaoyang Shi and
                  Laura Smith and
                  Jost Tobias Springenberg and
                  Kyle Stachowicz and
                  James Tanner and
                  Quan Vuong and
                  Homer Walke and
                  Anna Walling and
                  Haohuan Wang and
                  Lili Yu and
                  Ury Zhilinsky},
  title        = {{\(\pi\)}\({}_{\mbox{0.5}}\): a Vision-Language-Action Model with
                  Open-World Generalization},
  journal      = {CoRR},
  volume       = {abs/2504.16054},
  year         = {2025},
  url          = {https://doi.org/10.48550/arXiv.2504.16054},
  doi          = {10.48550/ARXIV.2504.16054},
  eprinttype    = {arXiv},
  eprint       = {2504.16054},
  bibsource    = {dblp computer science bibliography, https://dblp.org}
}

@inproceedings{DBLP:conf/corl/ZitkovichYXXXXW23,
  author       = {Brianna Zitkovich and
                  Tianhe Yu and
                  Sichun Xu and
                  Peng Xu and
                  Ted Xiao and
                  Fei Xia and
                  Jialin Wu and
                  Paul Wohlhart and
                  Stefan Welker and
                  Ayzaan Wahid and
                  Quan Vuong and
                  Vincent Vanhoucke and
                  Huong T. Tran and
                  Radu Soricut and
                  Anikait Singh and
                  Jaspiar Singh and
                  Pierre Sermanet and
                  Pannag R. Sanketi and
                  Grecia Salazar and
                  Michael S. Ryoo and
                  Krista Reymann and
                  Kanishka Rao and
                  Karl Pertsch and
                  Igor Mordatch and
                  Henryk Michalewski and
                  Yao Lu and
                  Sergey Levine and
                  Lisa Lee and
                  Tsang{-}Wei Edward Lee and
                  Isabel Leal and
                  Yuheng Kuang and
                  Dmitry Kalashnikov and
                  Ryan Julian and
                  Nikhil J. Joshi and
                  Alex Irpan and
                  Brian Ichter and
                  Jasmine Hsu and
                  Alexander Herzog and
                  Karol Hausman and
                  Keerthana Gopalakrishnan and
                  Chuyuan Fu and
                  Pete Florence and
                  Chelsea Finn and
                  Kumar Avinava Dubey and
                  Danny Driess and
                  Tianli Ding and
                  Krzysztof Marcin Choromanski and
                  Xi Chen and
                  Yevgen Chebotar and
                  Justice Carbajal and
                  Noah Brown and
                  Anthony Brohan and
                  Montserrat Gonzalez Arenas and
                  Kehang Han},
  editor       = {Jie Tan and
                  Marc Toussaint and
                  Kourosh Darvish},
  title        = {{RT-2:} Vision-Language-Action Models Transfer Web Knowledge to Robotic
                  Control},
  booktitle    = {Conference on Robot Learning, CoRL 2023, 6-9 November 2023, Atlanta,
                  GA, {USA}},
  series       = {Proceedings of Machine Learning Research},
  volume       = {229},
  pages        = {2165--2183},
  publisher    = {{PMLR}},
  year         = {2023},
  url          = {https://proceedings.mlr.press/v229/zitkovich23a.html},
  bibsource    = {dblp computer science bibliography, https://dblp.org}
}

@inproceedings{DBLP:conf/rss/GhoshWPBMDHK0LT24,
  author       = {Dibya Ghosh and
                  Homer Rich Walke and
                  Karl Pertsch and
                  Kevin Black and
                  Oier Mees and
                  Sudeep Dasari and
                  Joey Hejna and
                  Tobias Kreiman and
                  Charles Xu and
                  Jianlan Luo and
                  You Liang Tan and
                  Lawrence Yunliang Chen and
                  Quan Vuong and
                  Ted Xiao and
                  Pannag R. Sanketi and
                  Dorsa Sadigh and
                  Chelsea Finn and
                  Sergey Levine},
  editor       = {Dana Kulic and
                  Gentiane Venture and
                  Kostas E. Bekris and
                  Enrique Coronado},
  title        = {Octo: An Open-Source Generalist Robot Policy},
  booktitle    = {Robotics: Science and Systems XX, Delft, The Netherlands, July 15-19,
                  2024},
  year         = {2024},
  url          = {https://doi.org/10.15607/RSS.2024.XX.090},
  doi          = {10.15607/RSS.2024.XX.090},
  bibsource    = {dblp computer science bibliography, https://dblp.org}
}

@article{DBLP:journals/corr/abs-2405-14093,
  author       = {Yueen Ma and
                  Zixing Song and
                  Yuzheng Zhuang and
                  Jianye Hao and
                  Irwin King},
  title        = {A Survey on Vision-Language-Action Models for Embodied {AI}},
  journal      = {CoRR},
  volume       = {abs/2405.14093},
  year         = {2024},
  url          = {https://doi.org/10.48550/arXiv.2405.14093},
  doi          = {10.48550/ARXIV.2405.14093},
  eprinttype    = {arXiv},
  eprint       = {2405.14093},
  bibsource    = {dblp computer science bibliography, https://dblp.org}
}

@article{DBLP:journals/corr/abs-2502-05206,
  author       = {Xingjun Ma and
                  Yifeng Gao and
                  Yixu Wang and
                  Ruofan Wang and
                  Xin Wang and
                  Ye Sun and
                  Yifan Ding and
                  Hengyuan Xu and
                  Yunhao Chen and
                  Yunhan Zhao and
                  Hanxun Huang and
                  Yige Li and
                  Jiaming Zhang and
                  Xiang Zheng and
                  Yang Bai and
                  Zuxuan Wu and
                  Xipeng Qiu and
                  Jingfeng Zhang and
                  Yiming Li and
                  Jun Sun and
                  Cong Wang and
                  Jindong Gu and
                  Baoyuan Wu and
                  Siheng Chen and
                  Tianwei Zhang and
                  Yang Liu and
                  Mingming Gong and
                  Tongliang Liu and
                  Shirui Pan and
                  Cihang Xie and
                  Tianyu Pang and
                  Yinpeng Dong and
                  Ruoxi Jia and
                  Yang Zhang and
                  Shiqing Ma and
                  Xiangyu Zhang and
                  Neil Gong and
                  Chaowei Xiao and
                  Sarah M. Erfani and
                  Bo Li and
                  Masashi Sugiyama and
                  Dacheng Tao and
                  James Bailey and
                  Yu{-}Gang Jiang},
  title        = {Safety at Scale: {A} Comprehensive Survey of Large Model Safety},
  journal      = {CoRR},
  volume       = {abs/2502.05206},
  year         = {2025},
  url          = {https://doi.org/10.48550/arXiv.2502.05206},
  doi          = {10.48550/ARXIV.2502.05206},
  eprinttype    = {arXiv},
  eprint       = {2502.05206},
  bibsource    = {dblp computer science bibliography, https://dblp.org}
}

@article{DBLP:journals/access/NeupaneMFSMCPR24,
  author       = {Subash Neupane and
                  Shaswata Mitra and
                  Ivan A. Fernandez and
                  Swayamjit Saha and
                  Sudip Mittal and
                  Jingdao Chen and
                  Nisha Pillai and
                  Shahram Rahimi},
  title        = {Security Considerations in AI-Robotics: {A} Survey of Current Methods,
                  Challenges, and Opportunities},
  journal      = {{IEEE} Access},
  volume       = {12},
  pages        = {22072--22097},
  year         = {2024},
  url          = {https://doi.org/10.1109/ACCESS.2024.3363657},
  doi          = {10.1109/ACCESS.2024.3363657},
  bibsource    = {dblp computer science bibliography, https://dblp.org}
}

@article{DBLP:journals/corr/abs-2411-11683,
  author       = {Xianlong Wang and
                  Hewen Pan and
                  Hangtao Zhang and
                  Minghui Li and
                  Shengshan Hu and
                  Ziqi Zhou and
                  Lulu Xue and
                  Peijin Guo and
                  Yichen Wang and
                  Wei Wan and
                  Aishan Liu and
                  Leo Yu Zhang},
  title        = {TrojanRobot: Backdoor Attacks Against Robotic Manipulation in the
                  Physical World},
  journal      = {CoRR},
  volume       = {abs/2411.11683},
  year         = {2024},
  url          = {https://doi.org/10.48550/arXiv.2411.11683},
  doi          = {10.48550/ARXIV.2411.11683},
  eprinttype    = {arXiv},
  eprint       = {2411.11683},
  bibsource    = {dblp computer science bibliography, https://dblp.org}
}

@inproceedings{DBLP:conf/eccv/XuZ0FSCCWL20,
  author       = {Kaidi Xu and
                  Gaoyuan Zhang and
                  Sijia Liu and
                  Quanfu Fan and
                  Mengshu Sun and
                  Hongge Chen and
                  Pin{-}Yu Chen and
                  Yanzhi Wang and
                  Xue Lin},
  editor       = {Andrea Vedaldi and
                  Horst Bischof and
                  Thomas Brox and
                  Jan{-}Michael Frahm},
  title        = {Adversarial T-Shirt! Evading Person Detectors in a Physical World},
  booktitle    = {Computer Vision - {ECCV} 2020 - 16th European Conference, Glasgow,
                  UK, August 23-28, 2020, Proceedings, Part {V}},
  series       = {Lecture Notes in Computer Science},
  volume       = {12350},
  pages        = {665--681},
  publisher    = {Springer},
  year         = {2020},
  url          = {https://doi.org/10.1007/978-3-030-58558-7\_39},
  doi          = {10.1007/978-3-030-58558-7\_39},
  bibsource    = {dblp computer science bibliography, https://dblp.org}
}

@inproceedings{DBLP:conf/iclr/ZhangZ00YLXWHLG25,
  author       = {Hangtao Zhang and
                  Chenyu Zhu and
                  Xianlong Wang and
                  Ziqi Zhou and
                  Changgan Yin and
                  Minghui Li and
                  Lulu Xue and
                  Yichen Wang and
                  Shengshan Hu and
                  Aishan Liu and
                  Peijin Guo and
                  Leo Yu Zhang},
  title        = {BadRobot: Jailbreaking Embodied {LLM} Agents in the Physical World},
  booktitle    = {The Thirteenth International Conference on Learning Representations,
                  {ICLR} 2025, Singapore, April 24-28, 2025},
  publisher    = {OpenReview.net},
  year         = {2025},
  url          = {https://openreview.net/forum?id=ei3qCntB66},
  bibsource    = {dblp computer science bibliography, https://dblp.org}
}

@inproceedings{DBLP:conf/emnlp/LiSLZMQ21,
  author       = {Linyang Li and
                  Demin Song and
                  Xiaonan Li and
                  Jiehang Zeng and
                  Ruotian Ma and
                  Xipeng Qiu},
  editor       = {Marie{-}Francine Moens and
                  Xuanjing Huang and
                  Lucia Specia and
                  Scott Wen{-}tau Yih},
  title        = {Backdoor Attacks on Pre-trained Models by Layerwise Weight Poisoning},
  booktitle    = {Proceedings of the 2021 Conference on Empirical Methods in Natural
                  Language Processing, {EMNLP} 2021, Virtual Event / Punta Cana, Dominican
                  Republic, 7-11 November, 2021},
  pages        = {3023--3032},
  publisher    = {Association for Computational Linguistics},
  year         = {2021},
  url          = {https://doi.org/10.18653/v1/2021.emnlp-main.241},
  doi          = {10.18653/V1/2021.EMNLP-MAIN.241},
  bibsource    = {dblp computer science bibliography, https://dblp.org}
}

@article{DBLP:journals/corr/abs-2409-13174,
  author       = {Hao Cheng and
                  Erjia Xiao and
                  Chengyuan Yu and
                  Zhao Yao and
                  Jiahang Cao and
                  Qiang Zhang and
                  Jiaxu Wang and
                  Mengshu Sun and
                  Kaidi Xu and
                  Jindong Gu and
                  Renjing Xu},
  title        = {Manipulation Facing Threats: Evaluating Physical Vulnerabilities in
                  End-to-End Vision Language Action Models},
  journal      = {CoRR},
  volume       = {abs/2409.13174},
  year         = {2024},
  url          = {https://doi.org/10.48550/arXiv.2409.13174},
  doi          = {10.48550/ARXIV.2409.13174},
  eprinttype    = {arXiv},
  eprint       = {2409.13174},
  bibsource    = {dblp computer science bibliography, https://dblp.org}
}

@inproceedings{DBLP:conf/iclr/WuSKSFR25,
  author       = {Chen Henry Wu and
                  Rishi Rajesh Shah and
                  Jing Yu Koh and
                  Russ Salakhutdinov and
                  Daniel Fried and
                  Aditi Raghunathan},
  title        = {Dissecting Adversarial Robustness of Multimodal {LM} Agents},
  booktitle    = {The Thirteenth International Conference on Learning Representations,
                  {ICLR} 2025, Singapore, April 24-28, 2025},
  publisher    = {OpenReview.net},
  year         = {2025},
  url          = {https://openreview.net/forum?id=YauQYh2k1g},
  bibsource    = {dblp computer science bibliography, https://dblp.org}
}

@article{DBLP:journals/apin/XueWSZWL23,
  author       = {Mingfu Xue and
                  Xin Wang and
                  Shichang Sun and
                  Yushu Zhang and
                  Jian Wang and
                  Weiqiang Liu},
  title        = {Compression-resistant backdoor attack against deep neural networks},
  journal      = {Appl. Intell.},
  volume       = {53},
  number       = {17},
  pages        = {20402--20417},
  year         = {2023},
  url          = {https://doi.org/10.1007/s10489-023-04575-8},
  doi          = {10.1007/S10489-023-04575-8},
  bibsource    = {dblp computer science bibliography, https://dblp.org}
}

@inproceedings{DBLP:conf/nips/LiuYM22,
  author       = {Tian Yu Liu and
                  Yu Yang and
                  Baharan Mirzasoleiman},
  editor       = {Sanmi Koyejo and
                  S. Mohamed and
                  A. Agarwal and
                  Danielle Belgrave and
                  K. Cho and
                  A. Oh},
  title        = {Friendly Noise against Adversarial Noise: {A} Powerful Defense against
                  Data Poisoning Attack},
  booktitle    = {Advances in Neural Information Processing Systems 35: Annual Conference
                  on Neural Information Processing Systems 2022, NeurIPS 2022, New Orleans,
                  LA, USA, November 28 - December 9, 2022},
  year         = {2022},
  url          = {http://papers.nips.cc/paper\_files/paper/2022/hash/4e81308aa2eb8e2e4eccf122d4827af7-Abstract-Conference.html},
  bibsource    = {dblp computer science bibliography, https://dblp.org}
}

@inproceedings{DBLP:conf/corl/LiHGMPWFLSKL0F024,
  author       = {Xuanlin Li and
                  Kyle Hsu and
                  Jiayuan Gu and
                  Oier Mees and
                  Karl Pertsch and
                  Homer Rich Walke and
                  Chuyuan Fu and
                  Ishikaa Lunawat and
                  Isabel Sieh and
                  Sean Kirmani and
                  Sergey Levine and
                  Jiajun Wu and
                  Chelsea Finn and
                  Hao Su and
                  Quan Vuong and
                  Ted Xiao},
  editor       = {Pulkit Agrawal and
                  Oliver Kroemer and
                  Wolfram Burgard},
  title        = {Evaluating Real-World Robot Manipulation Policies in Simulation},
  booktitle    = {Conference on Robot Learning, 6-9 November 2024, Munich, Germany},
  series       = {Proceedings of Machine Learning Research},
  volume       = {270},
  pages        = {3705--3728},
  publisher    = {{PMLR}},
  year         = {2024},
  url          = {https://proceedings.mlr.press/v270/li25c.html},
  bibsource    = {dblp computer science bibliography, https://dblp.org}
}

@inproceedings{DBLP:conf/nips/LiuZGFLZS23,
  author       = {Bo Liu and
                  Yifeng Zhu and
                  Chongkai Gao and
                  Yihao Feng and
                  Qiang Liu and
                  Yuke Zhu and
                  Peter Stone},
  editor       = {Alice Oh and
                  Tristan Naumann and
                  Amir Globerson and
                  Kate Saenko and
                  Moritz Hardt and
                  Sergey Levine},
  title        = {{LIBERO:} Benchmarking Knowledge Transfer for Lifelong Robot Learning},
  booktitle    = {Advances in Neural Information Processing Systems 36: Annual Conference
                  on Neural Information Processing Systems 2023, NeurIPS 2023, New Orleans,
                  LA, USA, December 10 - 16, 2023},
  year         = {2023},
  url          = {http://papers.nips.cc/paper\_files/paper/2023/hash/8c3c666820ea055a77726d66fc7d447f-Abstract-Datasets\_and\_Benchmarks.html},
  bibsource    = {dblp computer science bibliography, https://dblp.org}
}

@inproceedings{DBLP:conf/aaai/0001HZZLL24,
  author       = {Zihan Guan and
                  Mengxuan Hu and
                  Zhongliang Zhou and
                  Jielu Zhang and
                  Sheng Li and
                  Ninghao Liu},
  editor       = {Michael J. Wooldridge and
                  Jennifer G. Dy and
                  Sriraam Natarajan},
  title        = {BadSAM: Exploring Security Vulnerabilities of {SAM} via Backdoor Attacks
                  (Student Abstract)},
  booktitle    = {Thirty-Eighth {AAAI} Conference on Artificial Intelligence, {AAAI}
                  2024, Thirty-Sixth Conference on Innovative Applications of Artificial
                  Intelligence, {IAAI} 2024, Fourteenth Symposium on Educational Advances
                  in Artificial Intelligence, {EAAI} 2014, February 20-27, 2024, Vancouver,
                  Canada},
  pages        = {23506--23507},
  publisher    = {{AAAI} Press},
  year         = {2024},
  url          = {https://doi.org/10.1609/aaai.v38i21.30448},
  doi          = {10.1609/AAAI.V38I21.30448},
  bibsource    = {dblp computer science bibliography, https://dblp.org}
}

@inproceedings{DBLP:conf/eccv/LyuPMLC24,
  author       = {Weimin Lyu and
                  Lu Pang and
                  Tengfei Ma and
                  Haibin Ling and
                  Chao Chen},
  editor       = {Ales Leonardis and
                  Elisa Ricci and
                  Stefan Roth and
                  Olga Russakovsky and
                  Torsten Sattler and
                  G{\"{u}}l Varol},
  title        = {TrojVLM: Backdoor Attack Against Vision Language Models},
  booktitle    = {Computer Vision - {ECCV} 2024 - 18th European Conference, Milan, Italy,
                  September 29-October 4, 2024, Proceedings, Part {LXV}},
  series       = {Lecture Notes in Computer Science},
  volume       = {15123},
  pages        = {467--483},
  publisher    = {Springer},
  year         = {2024},
  url          = {https://doi.org/10.1007/978-3-031-73650-6\_27},
  doi          = {10.1007/978-3-031-73650-6\_27},
  bibsource    = {dblp computer science bibliography, https://dblp.org}
}

@article{DBLP:journals/corr/abs-2411-19650,
  author       = {Qixiu Li and
                  Yaobo Liang and
                  Zeyu Wang and
                  Lin Luo and
                  Xi Chen and
                  Mozheng Liao and
                  Fangyun Wei and
                  Yu Deng and
                  Sicheng Xu and
                  Yizhong Zhang and
                  Xiaofan Wang and
                  Bei Liu and
                  Jianlong Fu and
                  Jianmin Bao and
                  Dong Chen and
                  Yuanchun Shi and
                  Jiaolong Yang and
                  Baining Guo},
  title        = {CogACT: {A} Foundational Vision-Language-Action Model for Synergizing
                  Cognition and Action in Robotic Manipulation},
  journal      = {CoRR},
  volume       = {abs/2411.19650},
  year         = {2024},
  url          = {https://doi.org/10.48550/arXiv.2411.19650},
  doi          = {10.48550/ARXIV.2411.19650},
  eprinttype    = {arXiv},
  eprint       = {2411.19650},
  bibsource    = {dblp computer science bibliography, https://dblp.org}
}

@inproceedings{DBLP:conf/iclr/ZhengLH0DKHY25,
  author       = {Ruijie Zheng and
                  Yongyuan Liang and
                  Shuaiyi Huang and
                  Jianfeng Gao and
                  Hal Daum{\'{e}} III and
                  Andrey Kolobov and
                  Furong Huang and
                  Jianwei Yang},
  title        = {TraceVLA: Visual Trace Prompting Enhances Spatial-Temporal Awareness
                  for Generalist Robotic Policies},
  booktitle    = {The Thirteenth International Conference on Learning Representations,
                  {ICLR} 2025, Singapore, April 24-28, 2025},
  publisher    = {OpenReview.net},
  year         = {2025},
  url          = {https://openreview.net/forum?id=b1CVu9l5GO},
  bibsource    = {dblp computer science bibliography, https://dblp.org}
}

@article{DBLP:journals/corr/abs-2506-01844,
  author       = {Mustafa Shukor and
                  Dana Aubakirova and
                  Francesco Capuano and
                  Pepijn Kooijmans and
                  Steven Palma and
                  Adil Zouitine and
                  Michel Aractingi and
                  Caroline Pascal and
                  Martino Russi and
                  Andr{\'{e}}s Marafioti and
                  Simon Alibert and
                  Matthieu Cord and
                  Thomas Wolf and
                  R{\'{e}}mi Cad{\`{e}}ne},
  title        = {SmolVLA: {A} Vision-Language-Action Model for Affordable and Efficient
                  Robotics},
  journal      = {CoRR},
  volume       = {abs/2506.01844},
  year         = {2025},
  url          = {https://doi.org/10.48550/arXiv.2506.01844},
  doi          = {10.48550/ARXIV.2506.01844},
  eprinttype    = {arXiv},
  eprint       = {2506.01844},
  bibsource    = {dblp computer science bibliography, https://dblp.org}
}

@article{DBLP:journals/corr/abs-2502-13175,
  author       = {Wenpeng Xing and
                  Minghao Li and
                  Mohan Li and
                  Meng Han},
  title        = {Towards Robust and Secure Embodied {AI:} {A} Survey on Vulnerabilities
                  and Attacks},
  journal      = {CoRR},
  volume       = {abs/2502.13175},
  year         = {2025},
  url          = {https://doi.org/10.48550/arXiv.2502.13175},
  doi          = {10.48550/ARXIV.2502.13175},
  eprinttype    = {arXiv},
  eprint       = {2502.13175},
  bibsource    = {dblp computer science bibliography, https://dblp.org}
}

@article{DBLP:journals/corr/abs-2503-03480,
  author       = {Borong Zhang and
                  Yuhao Zhang and
                  Jiaming Ji and
                  Yingshan Lei and
                  Josef Dai and
                  Yuanpei Chen and
                  Yaodong Yang},
  title        = {SafeVLA: Towards Safety Alignment of Vision-Language-Action Model
                  via Safe Reinforcement Learning},
  journal      = {CoRR},
  volume       = {abs/2503.03480},
  year         = {2025},
  url          = {https://doi.org/10.48550/arXiv.2503.03480},
  doi          = {10.48550/ARXIV.2503.03480},
  eprinttype    = {arXiv},
  eprint       = {2503.03480},
  bibsource    = {dblp computer science bibliography, https://dblp.org}
}

@article{DBLP:journals/corr/abs-2407-06886,
  author       = {Yang Liu and
                  Weixing Chen and
                  Yongjie Bai and
                  Guanbin Li and
                  Wen Gao and
                  Liang Lin},
  title        = {Aligning Cyber Space with Physical World: {A} Comprehensive Survey
                  on Embodied {AI}},
  journal      = {CoRR},
  volume       = {abs/2407.06886},
  year         = {2024},
  url          = {https://doi.org/10.48550/arXiv.2407.06886},
  doi          = {10.48550/ARXIV.2407.06886},
  eprinttype    = {arXiv},
  eprint       = {2407.06886},
  bibsource    = {dblp computer science bibliography, https://dblp.org}
}

@article{DBLP:journals/csur/LiuGC25,
  author       = {Huaping Liu and
                  Di Guo and
                  Angelo Cangelosi},
  title        = {Embodied Intelligence: {A} Synergy of Morphology, Action, Perception
                  and Learning},
  journal      = {{ACM} Comput. Surv.},
  volume       = {57},
  number       = {7},
  pages        = {186:1--186:36},
  year         = {2025},
  url          = {https://doi.org/10.1145/3717059},
  doi          = {10.1145/3717059},
  bibsource    = {dblp computer science bibliography, https://dblp.org}
}

@article{DBLP:journals/corr/abs-2507-00917,
  author       = {Xiaoxiao Long and
                  Qingrui Zhao and
                  Kaiwen Zhang and
                  Zihao Zhang and
                  Dingrui Wang and
                  Yumeng Liu and
                  Zhengjie Shu and
                  Yi Lu and
                  Shouzheng Wang and
                  Xinzhe Wei and
                  Wei Li and
                  Wei Yin and
                  Yao Yao and
                  Jia Pan and
                  Qiu Shen and
                  Ruigang Yang and
                  Xun Cao and
                  Qionghai Dai},
  title        = {A Survey: Learning Embodied Intelligence from Physical Simulators
                  and World Models},
  journal      = {CoRR},
  volume       = {abs/2507.00917},
  year         = {2025},
  url          = {https://doi.org/10.48550/arXiv.2507.00917},
  doi          = {10.48550/ARXIV.2507.00917},
  eprinttype    = {arXiv},
  eprint       = {2507.00917},
  bibsource    = {dblp computer science bibliography, https://dblp.org}
}

@article{DBLP:journals/corr/abs-2505-04769,
  author       = {Ranjan Sapkota and
                  Yang Cao and
                  Konstantinos I. Roumeliotis and
                  Manoj Karkee},
  title        = {Vision-Language-Action Models: Concepts, Progress, Applications and
                  Challenges},
  journal      = {CoRR},
  volume       = {abs/2505.04769},
  year         = {2025},
  url          = {https://doi.org/10.48550/arXiv.2505.04769},
  doi          = {10.48550/ARXIV.2505.04769},
  eprinttype    = {arXiv},
  eprint       = {2505.04769},
  bibsource    = {dblp computer science bibliography, https://dblp.org}
}

@inproceedings{DBLP:conf/corl/WalkeBZVZHHMKDL23,
  author       = {Homer Rich Walke and
                  Kevin Black and
                  Tony Z. Zhao and
                  Quan Vuong and
                  Chongyi Zheng and
                  Philippe Hansen{-}Estruch and
                  Andre Wang He and
                  Vivek Myers and
                  Moo Jin Kim and
                  Max Du and
                  Abraham Lee and
                  Kuan Fang and
                  Chelsea Finn and
                  Sergey Levine},
  editor       = {Jie Tan and
                  Marc Toussaint and
                  Kourosh Darvish},
  title        = {BridgeData {V2:} {A} Dataset for Robot Learning at Scale},
  booktitle    = {Conference on Robot Learning, CoRL 2023, 6-9 November 2023, Atlanta,
                  GA, {USA}},
  series       = {Proceedings of Machine Learning Research},
  volume       = {229},
  pages        = {1723--1736},
  publisher    = {{PMLR}},
  year         = {2023},
  url          = {https://proceedings.mlr.press/v229/walke23a.html},
  bibsource    = {dblp computer science bibliography, https://dblp.org}
}

@article{DBLP:journals/corr/abs-2506-07214,
  author       = {Zhiyuan Zhong and
                  Zhen Sun and
                  Yepang Liu and
                  Xinlei He and
                  Guanhong Tao},
  title        = {Backdoor Attack on Vision Language Models with Stealthy Semantic Manipulation},
  journal      = {CoRR},
  volume       = {abs/2506.07214},
  year         = {2025},
  url          = {https://doi.org/10.48550/arXiv.2506.07214},
  doi          = {10.48550/ARXIV.2506.07214},
  eprinttype    = {arXiv},
  eprint       = {2506.07214},
  bibsource    = {dblp computer science bibliography, https://dblp.org}
}

@article{DBLP:journals/corr/abs-2505-06413,
  author       = {Ming Liu and
                  Siyuan Liang and
                  Koushik Howlader and
                  Liwen Wang and
                  Dacheng Tao and
                  Wensheng Zhang},
  title        = {Natural Reflection Backdoor Attack on Vision Language Model for Autonomous
                  Driving},
  journal      = {CoRR},
  volume       = {abs/2505.06413},
  year         = {2025},
  url          = {https://doi.org/10.48550/arXiv.2505.06413},
  doi          = {10.48550/ARXIV.2505.06413},
  eprinttype    = {arXiv},
  eprint       = {2505.06413},
  bibsource    = {dblp computer science bibliography, https://dblp.org}
}

@inproceedings{DBLP:conf/cvpr/LiangLPDLZCT25,
  author       = {Siyuan Liang and
                  Jiawei Liang and
                  Tianyu Pang and
                  Chao Du and
                  Aishan Liu and
                  Mingli Zhu and
                  Xiaochun Cao and
                  Dacheng Tao},
  title        = {Revisiting Backdoor Attacks against Large Vision-Language Models from
                  Domain Shift},
  booktitle    = {{IEEE/CVF} Conference on Computer Vision and Pattern Recognition,
                  {CVPR} 2025, Nashville, TN, USA, June 11-15, 2025},
  pages        = {9477--9486},
  publisher    = {Computer Vision Foundation / {IEEE}},
  year         = {2025},
  url          = {https://openaccess.thecvf.com/content/CVPR2025/html/Liang\_Revisiting\_Backdoor\_Attacks\_against\_Large\_Vision-Language\_Models\_from\_Domain\_Shift\_CVPR\_2025\_paper.html},
  doi          = {10.1109/CVPR52734.2025.00885},
  bibsource    = {dblp computer science bibliography, https://dblp.org}
}

@article{DBLP:journals/corr/abs-2502-19269,
  author       = {Jiawei Kong and
                  Hao Fang and
                  Sihang Guo and
                  Chenxi Qing and
                  Bin Chen and
                  Bin Wang and
                  Shu{-}Tao Xia},
  title        = {Neural Antidote: Class-Wise Prompt Tuning for Purifying Backdoors
                  in Pre-trained Vision-Language Models},
  journal      = {CoRR},
  volume       = {abs/2502.19269},
  year         = {2025},
  url          = {https://doi.org/10.48550/arXiv.2502.19269},
  doi          = {10.48550/ARXIV.2502.19269},
  eprinttype    = {arXiv},
  eprint       = {2502.19269},
  bibsource    = {dblp computer science bibliography, https://dblp.org}
}

@article{DBLP:journals/corr/abs-2506-05401,
  author       = {Yuan Xun and
                  Siyuan Liang and
                  Xiaojun Jia and
                  Xinwei Liu and
                  Xiaochun Cao},
  title        = {Robust Anti-Backdoor Instruction Tuning in LVLMs},
  journal      = {CoRR},
  volume       = {abs/2506.05401},
  year         = {2025},
  url          = {https://doi.org/10.48550/arXiv.2506.05401},
  doi          = {10.48550/ARXIV.2506.05401},
  eprinttype    = {arXiv},
  eprint       = {2506.05401},
  bibsource    = {dblp computer science bibliography, https://dblp.org}
}

@article{vuong2023open,
author={Vuong, Quan and Levine, Sergey and Walke, Homer Rich and Pertsch, Karl and Singh, Anikait and Doshi, Ria and Xu, Charles and Luo, Jianlan and Tan, Liam and Shah, Dhruv and others},
journal={CoRL},
title={Open x-embodiment: Robotic learning datasets and rt-x models},
year={2023}
}

@misc{cadene2024lerobot,
    author = {Cadene, Remi and Alibert, Simon and Soare, Alexander and Gallouedec, Quentin and Zouitine, Adil and Palma, Steven and Kooijmans, Pepijn and Aractingi, Michel and Shukor, Mustafa and Aubakirova, Dana and Russi, Martino and Capuano, Francesco and Pascale, Caroline and Choghari, Jade and Moss, Jess and Wolf, Thomas},
    title = {LeRobot: State-of-the-art Machine Learning for Real-World Robotics in Pytorch},
    howpublished = "\url{https://github.com/huggingface/lerobot}",
    year = {2024}
}

@inproceedings{DBLP:conf/icml/Kornblith0LH19,
  author       = {Simon Kornblith and
                  Mohammad Norouzi and
                  Honglak Lee and
                  Geoffrey E. Hinton},
  editor       = {Kamalika Chaudhuri and
                  Ruslan Salakhutdinov},
  title        = {Similarity of Neural Network Representations Revisited},
  booktitle    = {Proceedings of the 36th International Conference on Machine Learning,
                  {ICML} 2019, 9-15 June 2019, Long Beach, California, {USA}},
  series       = {Proceedings of Machine Learning Research},
  volume       = {97},
  pages        = {3519--3529},
  publisher    = {{PMLR}},
  year         = {2019},
  url          = {http://proceedings.mlr.press/v97/kornblith19a.html},
  bibsource    = {dblp computer science bibliography, https://dblp.org}
}

@article{DBLP:journals/pacmse/WangZSHSM25,
  author       = {Zhijie Wang and
                  Zhehua Zhou and
                  Jiayang Song and
                  Yuheng Huang and
                  Zhan Shu and
                  Lei Ma},
  title        = {VLATest: Testing and Evaluating Vision-Language-Action Models for
                  Robotic Manipulation},
  journal      = {Proc. {ACM} Softw. Eng.},
  volume       = {2},
  number       = {{FSE}},
  pages        = {1615--1638},
  year         = {2025},
  url          = {https://doi.org/10.1145/3729343},
  doi          = {10.1145/3729343},
  bibsource    = {dblp computer science bibliography, https://dblp.org}
}

@inproceedings{DBLP:conf/cvpr/WuZX0Y25,
  author       = {Zhenyu Wu and
                  Yuheng Zhou and
                  Xiuwei Xu and
                  Ziwei Wang and
                  Haibin Yan},
  title        = {MoManipVLA: Transferring Vision-language-action Models for General
                  Mobile Manipulation},
  booktitle    = {{IEEE/CVF} Conference on Computer Vision and Pattern Recognition,
                  {CVPR} 2025, Nashville, TN, USA, June 11-15, 2025},
  pages        = {1714--1723},
  publisher    = {Computer Vision Foundation / {IEEE}},
  year         = {2025},
  url          = {https://openaccess.thecvf.com/content/CVPR2025/html/Wu\_MoManipVLA\_Transferring\_Vision-language-action\_Models\_for\_General\_Mobile\_Manipulation\_CVPR\_2025\_paper.html},
  doi          = {10.1109/CVPR52734.2025.00167},
  bibsource    = {dblp computer science bibliography, https://dblp.org}
}

@inproceedings{DBLP:conf/cvpr/ZhaoLKFZWLMHFHL25,
  author       = {Qingqing Zhao and
                  Yao Lu and
                  Moo Jin Kim and
                  Zipeng Fu and
                  Zhuoyang Zhang and
                  Yecheng Wu and
                  Zhaoshuo Li and
                  Qianli Ma and
                  Song Han and
                  Chelsea Finn and
                  Ankur Handa and
                  Tsung{-}Yi Lin and
                  Gordon Wetzstein and
                  Ming{-}Yu Liu and
                  Donglai Xiang},
  title        = {CoT-VLA: Visual Chain-of-Thought Reasoning for Vision-Language-Action
                  Models},
  booktitle    = {{IEEE/CVF} Conference on Computer Vision and Pattern Recognition,
                  {CVPR} 2025, Nashville, TN, USA, June 11-15, 2025},
  pages        = {1702--1713},
  publisher    = {Computer Vision Foundation / {IEEE}},
  year         = {2025},
  url          = {https://openaccess.thecvf.com/content/CVPR2025/html/Zhao\_CoT-VLA\_Visual\_Chain-of-Thought\_Reasoning\_for\_Vision-Language-Action\_Models\_CVPR\_2025\_paper.html},
  doi          = {10.1109/CVPR52734.2025.00166},
  bibsource    = {dblp computer science bibliography, https://dblp.org}
}

@inproceedings{DBLP:conf/hri/SautenkovYLMTAC25,
  author       = {Oleg Sautenkov and
                  Yasheerah Yaqoot and
                  Artem Lykov and
                  Muhammad Ahsan Mustafa and
                  Grik Tadevosyan and
                  Aibek Akhmetkazy and
                  Miguel Altamirano Cabrera and
                  Mikhail Martynov and
                  Sausar Karaf and
                  Dzmitry Tsetserukou},
  title        = {{UAV-VLA:} Vision-Language-Action System for Large Scale Aerial Mission
                  Generation},
  booktitle    = {20th {ACM/IEEE} International Conference on Human-Robot Interaction,
                  {HRI} 2025, Melbourne, Australia, March 4-6, 2025},
  pages        = {1588--1592},
  publisher    = {{IEEE}},
  year         = {2025},
  url          = {https://doi.org/10.1109/HRI61500.2025.10974117},
  doi          = {10.1109/HRI61500.2025.10974117},
  bibsource    = {dblp computer science bibliography, https://dblp.org}
}

@inproceedings{DBLP:conf/wacv/AraiMSWY0Y25,
  author       = {Hidehisa Arai and
                  Keita Miwa and
                  Kento Sasaki and
                  Kohei Watanabe and
                  Yu Yamaguchi and
                  Shunsuke Aoki and
                  Issei Yamamoto},
  title        = {CoVLA: Comprehensive Vision-Language-Action Dataset for Autonomous
                  Driving},
  booktitle    = {{IEEE/CVF} Winter Conference on Applications of Computer Vision, {WACV}
                  2025, Tucson, AZ, USA, February 26 - March 6, 2025},
  pages        = {1933--1943},
  publisher    = {{IEEE}},
  year         = {2025},
  url          = {https://doi.org/10.1109/WACV61041.2025.00195},
  doi          = {10.1109/WACV61041.2025.00195},
  bibsource    = {dblp computer science bibliography, https://dblp.org}
}

@article{DBLP:journals/caaitrit/WangNLGDZ25,
  author       = {Sicheng Wang and
                  Milutin N. Nikolic and
                  Tin Lun Lam and
                  Qing Gao and
                  Runwei Ding and
                  Tianwei Zhang},
  title        = {Robot Manipulation Based on Embodied Visual Perception: {A} Survey},
  journal      = {{CAAI} Trans. Intell. Technol.},
  volume       = {10},
  number       = {4},
  pages        = {945--958},
  year         = {2025},
  url          = {https://doi.org/10.1049/cit2.70022},
  doi          = {10.1049/CIT2.70022},
  bibsource    = {dblp computer science bibliography, https://dblp.org}
}

@article{DBLP:journals/corr/abs-2508-13901,
  author       = {Yihao Lu and
                  Hao Tang},
  title        = {Multimodal Data Storage and Retrieval for Embodied {AI:} {A} Survey},
  journal      = {CoRR},
  volume       = {abs/2508.13901},
  year         = {2025},
  url          = {https://doi.org/10.48550/arXiv.2508.13901},
  doi          = {10.48550/ARXIV.2508.13901},
  eprinttype    = {arXiv},
  eprint       = {2508.13901},
  bibsource    = {dblp computer science bibliography, https://dblp.org}
}

@article{DBLP:journals/corr/abs-2402-02385,
  author       = {Zhiyuan Xu and
                  Kun Wu and
                  Junjie Wen and
                  Jinming Li and
                  Ning Liu and
                  Zhengping Che and
                  Jian Tang},
  title        = {A Survey on Robotics with Foundation Models: toward Embodied {AI}},
  journal      = {CoRR},
  volume       = {abs/2402.02385},
  year         = {2024},
  url          = {https://doi.org/10.48550/arXiv.2402.02385},
  doi          = {10.48550/ARXIV.2402.02385},
  eprinttype    = {arXiv},
  eprint       = {2402.02385},
  bibsource    = {dblp computer science bibliography, https://dblp.org}
}
}

\end{document}